\documentclass{article} %
\usepackage{iclr2027_conference,times}

\usepackage[utf8]{inputenc}
\usepackage[T1]{fontenc}
\usepackage{amsmath,amssymb,amsthm}
\usepackage{graphicx}
\usepackage{tikz}
\usetikzlibrary{positioning,arrows.meta,fit,backgrounds,calc,shapes.geometric,decorations.pathreplacing}
\usepackage{booktabs}        %
\usepackage{caption}         %
\usepackage{array}
\newcolumntype{L}[1]{>{\raggedright\arraybackslash}p{#1}}
\newlength{\figstrengthht} %
\usepackage{microtype}
\usepackage{xcolor}
\usepackage[colorlinks=true,linkcolor=blue,citecolor=blue,urlcolor=blue]{hyperref}
\usepackage{url}
\usepackage{enumitem}
\usepackage[capitalize,noabbrev]{cleveref}

\newcommand{\poke}{Pok\'emon}

\title{PokaiTrainer: Scaling Equilibrium Search to\\Competitive Pok\'emon VGC}

\author{Max Yu \\ Independent Researcher}

\iclrfinalcopy %
\def\arxivbuild{} %

\begin{document}

\maketitle
\lhead{Preprint}

\begin{abstract}
Decision-time equilibrium search carried poker to superhuman play, but it has so far relied on tractable subgames: a handful of actions per decision, chance confined to card deals, one player moving at a time.
Competitive \poke\ in its official doubles format (VGC) breaks all three assumptions at once.
Both players act simultaneously from joint menus in the hundreds, each joint action resolves to hundreds of stochastic outcomes, and the opponent's reserves and stat allocations are hidden.
No prior \poke\ agent performs equilibrium search, and whether it scales to this regime was open; we show that it does, and report what it took.
PokaiEngine, our Rust battle engine, enumerates a joint action's full weighted outcome distribution in one pass, at ${\sim}99\%$ parity with \poke\ Showdown and a fraction of the cost of sampling it.
PokaiTrainer adapts Student of Games to this scale and trains it by self-play over hundreds of human teams. Each decision is solved by counterfactual regret minimization as a Bayesian matrix game over public belief states, subgames grow under an explicit compute budget, and value targets are harvested from the interior of every solve and grounded by realized outcomes.
The strength is in the search. The network's policy alone loses even to a shallow heuristic search.
PokaiTrainer is, to our knowledge, the first VGC agent rated on the live Showdown ladder. Under open team sheets it wins 59\% of 150 best-of-three sets against a human field averaging ${\sim}1320$ Elo, holds a 1350--1400 Elo band, and at its peak reached 1492 Elo, entering the format's top 500.
\end{abstract}

\section{Introduction}
\label{sec:intro}

Equilibrium search has produced landmark results in imperfect-information games, carrying poker from competitive to superhuman play \citep{zinkevich2007cfr,moravcik2017deepstack,brown2018libratus,brown2019pluribus}.
Its modern form, exemplified by ReBeL \citep{brown2020rebel} and Student of Games \citep{schmid2023student}, solves a depth-limited subgame rooted at the current public belief state, with a learned value function at the leaves and self-play training.
The recipe has so far relied on tractable subgames: poker offers a handful of actions per decision and confines chance to card deals, so ReBeL can enumerate its subgames exhaustively.
Even Student of Games, growing its subgames incrementally, branches on a single player's move at a time.
Equilibrium search also reached human level in Diplomacy \citep{gray2020diplomacy,fair2022diplomacy}, whose turns are simultaneous, but the game is deterministic and the search one step deep over a few dozen sampled actions.

Competitive \poke\ Battling, in its official doubles format (VGC), breaks that tractability on every axis.
Turns are simultaneous, with hundreds to thousands of joint actions, and each joint action resolves to hundreds of possible stochastic outcomes.
Prior \poke\ agents sidestep this: whether heuristic or search-based, LLM-driven, or learned by imitation and RL \citep{lee2017showdownai,hu2024pokellmon,karten2025pokechamp,grigsby2025metamon,angliss2025vgcbench}, none performs equilibrium search at decision time (\cref{sec:related}).
Whether equilibrium search scales to this regime has been an open question, and we report what answering it took.

To handle the outcome stochasticity of \poke, we built \textbf{PokaiEngine}, a Rust battle engine that enumerates the chance outcomes of a joint action, returning its weighted outcome distribution in one pass.
It reaches ${\sim}99\%$ behavioral parity with \poke\ Showdown \citep{showdown}, simulates single timelines $26\times$ faster, and recovers an outcome distribution more accurately than 256 Showdown replays at $1/500$th of their cost (\cref{sec:engine}).

We then train \textbf{PokaiTrainer}, a search agent in the Student of Games~\citep{schmid2023student} mold: CFR-based subgame solving over public belief states with value-network leaves, the subgame grown incrementally under an explicit compute budget, trained by self-play.
Decisions are solved as simultaneous-move Bayesian matrix games with enumerated chance nodes; expansion is priced by compute cost rather than node count; hidden stat spreads are inferred by re-branching observed turns through the engine itself; and supervision is harvested from the interior of every solve, grounded by realized outcomes, and extended by hypothetical augmentation to positions self-play under-visits.

The trained agent wins 59\% of 150 best-of-three sets on the live Showdown ladder against a human field averaging ${\sim}1320$ Elo, holding a 1350--1400 Elo band and at its peak entering the format's top 500 (\cref{sec:exp-ladder}).
Search carries the strength, as the policy network alone loses even to a shallow material-heuristic search. The equilibrium makes the agent robust to an informed best response (\cref{sec:exp-selfplay}).
Training returns are now steeply diminishing, and the defect that persists is the value network's early-game optimism, which deeper solves mask rather than fix.

We evaluate under Open Team Sheets and leave closed sheets, where the opponent's unrevealed sets must be inferred, as future work.
We claim no algorithmic novelty beyond the adaptations above. The contributions are the system, the empirical findings on what this scale demands, and a mapped frontier of what remains hard.
We intend to release the engine and trainer with the live-ladder client removed, and to withhold checkpoints while their format is in play (\hyperref[sec:ethics]{Ethics Statement}).

\section{Background}
\label{sec:background}

\subsection{Competitive \poke\ VGC}
\label{sec:background-vgc}

VGC is the official competitive \poke\ format: a \textbf{two-player zero-sum (2p0s)} game in which each player brings a team of 6 customized \poke.
A battle opens with \textit{team preview}: both teams are revealed, then each player secretly selects 4 of their 6, leading with two on the field and holding two in reserve.
Each turn both players simultaneously choose a move or a switch for each on-field \poke; actions resolve in order of priority and speed, switches first, and moves deal damage, build indirect advantage (stat boosts, disruption, field effects), or pivot the user out mid-turn for a replacement.
Fainted \poke\ are replaced from reserves at the end of the turn, and a player with none left loses.
The following properties make this game challenging:

\paragraph{Simultaneous moves.}
Both players commit before either action resolves, so every turn is a matrix game with genuine rock--paper--scissors structure: \textit{Protect} blocks attacks aimed at the user but wastes the action if the opponent targets elsewhere, and switching may dodge a predicted hit but invites a punish.
Deterministic play is exploitable, so optimal play requires mixing.

\paragraph{Imperfect information.}
Each \poke\ is configured at team-building time: a species, an ability, a held item, four moves, and a stat allocation (a nature and a budget of effort values) that tunes its speed, power, and bulk.
\emph{Closed Team Sheets} hide everything beyond the six species; official tournament play and the Showdown best-of-three (Bo3) ladder we evaluate on use \emph{Open Team Sheets} (OTS), which reveal everything but the stat allocations.
OTS still leaves three sources of hidden information: the stat spreads, which shift speed order and knock-out thresholds; which two of the four non-leads were brought; and the opponent's live commitment on every simultaneous turn.

\paragraph{Stochasticity.}
Even a basic damaging move draws damage from 16 uniform multipliers and can miss, critically strike, or trigger a secondary effect. Up to four moves resolve per turn, compounding to the outcome counts of \cref{tab:game-comparison} even in the low-variance lines competitive play favors.

\begin{table}[t]
\centering\small
\setlength{\tabcolsep}{4pt}%
\begin{tabular}{@{}l >{\raggedright\arraybackslash}p{0.10\linewidth} >{\raggedright\arraybackslash}p{0.12\linewidth} >{\raggedright\arraybackslash}p{0.10\linewidth} >{\raggedright\arraybackslash}p{0.145\linewidth} >{\raggedright\arraybackslash}p{0.22\linewidth}@{}}
\toprule
 & Chess / Go & HUNL poker & Scotland Yard & No-press Diplomacy & VGC doubles \\
\midrule
Simultaneous actors & --- & --- & --- & 7 powers & 2 players $\times$ 2 slots \\
\addlinespace[2pt]
Hidden information & --- & ${\sim}10^3$ hands & $\le$199 stations & --- & 15 brings $\times$ ${\sim}10^7$ spreads per \poke \\
\addlinespace[2pt]
Chance & --- & card deals only & --- & --- & $10^2$--$10^3$ outcomes per joint action \\
\addlinespace[2pt]
Actions per decision & $\sim$35 / $\sim$250 & a few bet sizes & $\sim$10 & up to ${\sim}10^{24}$ per power; a few dozen searched & $\sim 10^2$ per player, $\sim 10^4$ joint \\
\addlinespace[2pt]
Horizon & $\sim$80 / $\sim$150 plies & $\leq$4 bet rounds & 24 rounds & open-ended & 5--20 turns \\
\bottomrule
\end{tabular}
\caption{VGC compared to the domains decision-time equilibrium search has mastered: the four of Student of Games \citep{schmid2023student} and no-press Diplomacy \citep{gray2020diplomacy}. The ${\sim}10^7$ spread grid matters less than its size suggests (\cref{sec:method-spreads}).} %
\label{tab:game-comparison}
\end{table}

\paragraph{Mechanical density.}
Each format admits hundreds of species, moves, abilities, and items, plus a once-per-battle gimmick (Mega Evolution in Reg-MB). Simulation is solved \citep{showdown}, so the burden falls on learning, where interaction-specific knowledge competes with strategic knowledge for capacity, motivating search that computes it on demand.

\subsection{Equilibrium Search in Imperfect-Information Games}
\label{sec:background-search}
The properties above make a VGC battle a two-player zero-sum extensive-form game with simultaneous moves, imperfect information, and chance.
Strong human players adapt over a few games and beat the VGC-Bench agent \citep{angliss2025vgcbench}, and in our experience the four an agent brings is the first thing opponents read in a best-of-three.
In this class a Nash equilibrium is unexploitable, so we target approximate equilibrium play with the toolkit developed in poker: counterfactual regret minimization (CFR) \citep{zinkevich2007cfr,bowling2015hulhe}, combined with decision-time search that solves a \emph{subgame} rooted at the current \emph{public belief state} (PBS)---the public history plus each player's distribution over private states---\citep{brown2017safe,brown2018libratus,moravcik2017deepstack,brown2019pluribus}, cut off at a fixed depth by a learned \emph{counterfactual value network} returning one leaf value per private state \citep{moravcik2017deepstack}.
ReBeL \citep{brown2020rebel} closed the training loop, learning the value network from the subgames solved during self-play.
Student of Games (SoG) \citep{schmid2023student} extends the recipe to public trees too large to build up front---growing-tree CFR (GT-CFR) interleaves regret updates with expansions chosen by a walk mixing PUCT with CFR's average policy---and supplements main-line targets with re-solved leaf \emph{queries}.
As \cref{tab:game-comparison} shows, each of these branches on one player's move at a time; \cref{sec:method} is what the recipe needs when it does not.

\section{The PokaiEngine}
\label{sec:engine}

A standard \poke\ engine, both the games themselves and Showdown \citep{showdown}, advances a battle down a single timeline of chance rolls.
What matters about a joint action for evaluating a decision, however, is its outcome distribution, which a one-path engine exposes only by replaying the turn many times, and since chance compounds across the moves of a turn, even moderately likely outcomes get missed at practical sample counts. %

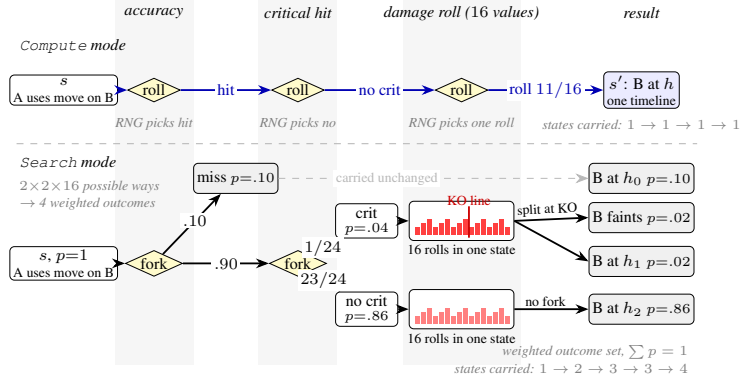
\begin{figure}[t]
\centering
\resizebox{0.7\textwidth}{!}{\begin{tikzpicture}[
  font=\scriptsize, >={Stealth[length=2mm]},
  st/.style={draw, rounded corners=2pt, fill=white, minimum width=1.15cm, minimum height=0.46cm, inner sep=1.5pt, align=center},
  term/.style={st, fill=gray!12},
  chance/.style={draw, diamond, aspect=2.2, inner sep=0.4pt, fill=yellow!25, align=center},
  lab/.style={font=\scriptsize\itshape, inner sep=1pt},
  w/.style={font=\scriptsize, inner sep=1pt, fill=white},
  tn/.style={font=\tiny, inner sep=1pt},
  comp/.style={->, thick, blue!70!black},
  srch/.style={->, thick},
  dead/.style={->, gray!55, dashed},
]
\begin{scope}[on background layer]
  \fill[gray!7] (1.5,0.42) rectangle (2.7,-5.2);
  \fill[gray!7] (3.7,0.42) rectangle (4.9,-5.2);
  \fill[gray!7] (5.9,0.42) rectangle (7.7,-5.2);
\end{scope}
\node[lab] at (2.1,0.18) {accuracy};
\node[lab] at (4.3,0.18) {critical hit};
\node[lab] at (6.8,0.18) {damage roll ($16$ values)};
\node[lab] at (9.55,0.18) {result};
\node[lab, anchor=west] at (0,-0.32) {\texttt{Compute} mode};
\node[st] (c0) at (0.72,-1.0) {$s$\\[-1pt]\tiny A uses move on B};
\node[chance] (c1) at (2.1,-1.0) {roll};
\node[chance] (c2) at (4.3,-1.0) {roll};
\node[chance] (c3) at (6.8,-1.0) {roll};
\node[term, fill=blue!8] (c4) at (9.55,-1.0) {$s'$: B at $h$\\[-1pt]\tiny one timeline};
\draw[comp] (c0) -- (c1);
\draw[comp] (c1) -- node[w] {hit} (c2);
\draw[comp] (c2) -- node[w] {no crit} (c3);
\draw[comp] (c3) -- node[w] {roll $11/16$} (c4);
\foreach \x/\t in {2.1/{RNG picks hit}, 4.3/{RNG picks no}, 6.8/{RNG picks one roll}} \node[lab, gray] at (\x,-1.52) {\tiny \t};
\node[lab, gray] at (9.55,-1.52) {\tiny states carried: $1\to1\to1\to1$};
\draw[gray!60, dashed] (0,-1.82) -- (10.4,-1.82);
\node[lab, anchor=west] at (0,-2.1) {\texttt{Search} mode};
\node[lab, gray, anchor=west] at (0,-2.48) {\tiny $2{\times}2{\times}16$ possible ways};
\node[lab, gray, anchor=west] at (0,-2.74) {\tiny $\to$ 4 weighted outcomes};
\node[st] (s0) at (0.72,-3.65) {$s$, $p{=}1$\\[-1pt]\tiny A uses move on B};
\node[chance] (s1) at (2.1,-3.65) {fork};
\draw[srch] (s0) -- (s1);
\node[term] (smiss) at (3.35,-2.35) {miss \tiny $p{=}.10$};
\draw[srch] (s1) -- node[w, pos=0.55] {$.10$} (smiss);
\node[term] (omiss) at (9.55,-2.35) {B at $h_0$ \tiny $p{=}.10$};
\draw[dead] (smiss) -- node[w, pos=0.35] {\tiny carried unchanged} (omiss);
\node[chance] (s2) at (4.3,-3.65) {fork};
\draw[srch] (s1) -- node[w] {$.90$} (s2);
\node[st, minimum width=0.95cm] (scrit) at (5.35,-3.0) {crit\\[-2pt]\tiny $p{=}.04$};
\node[st, minimum width=0.95cm] (snocrit) at (5.35,-4.35) {no crit\\[-2pt]\tiny $p{=}.86$};
\draw[srch] (s2) -- node[w, pos=0.45] {$1/24$} (scrit);
\draw[srch] (s2) -- node[w, pos=0.45] {$23/24$} (snocrit);
\node[st, minimum width=1.6cm, minimum height=0.6cm] (hcrit) at (6.8,-3.0) {};
\node[st, minimum width=1.6cm, minimum height=0.6cm] (hnocrit) at (6.8,-4.35) {};
\foreach \i in {0,...,15} {
  \fill[red!75] ($(hcrit.south west)+(0.09+\i*0.092,0.09)$) rectangle ++(0.074,{0.11+0.065*mod(\i,3)});
  \fill[red!50] ($(hnocrit.south west)+(0.09+\i*0.092,0.09)$) rectangle ++(0.074,{0.11+0.065*mod(\i,3)});
}
\draw[red!80!black, thick] ($(hcrit.south west)+(0.92,0.05)$) -- ++(0,0.48) node[above, font=\tiny, inner sep=1pt, yshift=-1pt] {KO line};
\node[tn, anchor=north] at ($(hcrit.south)+(0,-0.02)$) {16 rolls in one state};
\node[tn, anchor=north] at ($(hnocrit.south)+(0,-0.02)$) {16 rolls in one state};
\draw[srch] (scrit) -- (hcrit);
\draw[srch] (snocrit) -- (hnocrit);
\node[term] (oko) at (9.55,-2.95) {B faints \tiny $p{=}.02$};
\node[term] (ocrit) at (9.55,-3.62) {B at $h_1$ \tiny $p{=}.02$};
\node[term] (onocrit) at (9.55,-4.35) {B at $h_2$ \tiny $p{=}.86$};
\draw[srch] (hcrit.east) -- node[w, pos=0.45, above, sloped, fill=none] {\tiny split at KO} (oko.west);
\draw[srch] (hcrit.east) -- (ocrit.west);
\draw[srch] (hnocrit.east) -- node[w, pos=0.42, above, fill=none] {\tiny no fork} (onocrit.west);
\node[lab, gray, anchor=east] at (10.3,-5.0) {\tiny weighted outcome set, $\sum p = 1$};
\node[lab, gray, anchor=east] at (10.3,-5.26) {\tiny states carried: $1\to2\to3\to3\to4$};
\end{tikzpicture}}
\caption{One attack's three chance events under the two engine modes.
\texttt{Compute} rolls each event and continues a single timeline.
\texttt{Search} forks the state and carries every branch, holding the 16 damage rolls inside one state as an HP distribution split only where it straddles a knock-out.}
\label{fig:engine-modes}
\end{figure}

PokaiEngine, our Rust battle engine, instead supports chance-based forking natively.
The same code runs in two modes.
At each chance event, \texttt{Compute} mode rolls the RNG and continues down one timeline, while \texttt{Search} mode forks the state and carries every branch forward with its probability weight, as \cref{fig:engine-modes} traces for one attack.
Resolving a turn in \texttt{Search} mode therefore returns the joint action's full weighted outcome distribution in a single pass, with no sampling variance.\footnote{In the reported run, branches below a 2\% probability floor are pruned and at most eight are kept per joint action, retaining 97\% of the outcome mass on the probe of \cref{tab:outcome-coverage}, and identical states are merged.}

\subsection{Optimizations}
Forked naively, a turn multiplies out fast (16 damage rolls, an accuracy, a critical-hit, and a secondary-effect roll per move, up to four moves).
Two optimizations keep enumeration cheap (details in \cref{app:engine-opt}).
\emph{Representing outcomes compactly}: branching that changes numbers without changing behavior stays inside a single state, as \cref{fig:engine-modes} shows for HP, and hidden stat spreads can optionally ride the same way, assuming \textbf{stat spreads of different \poke\ are independent}.
A fork is spent only where outcomes diverge in kind rather than degree, such as a damage roll that straddles a knock-out or an uncertain speed order.
\emph{Sharing work across branches}: the closely related forked states of a turn advance together, batched on the attributes move resolution can share.

\begin{table}[t]
\centering
\small
\begin{tabular}{lrrr}
\toprule
Method & Mean time (ms) & Mass covered & Mass misallocated \\
\midrule
\texttt{Search}, exhaustive & 9.9 & 100\% & 0\% (reference) \\
\texttt{Search}, 32-branch cap & 1.0 & 99.5\% & 0.7\% \\
\texttt{Compute}\,/\,Showdown sampling, $N{=}16$ & 1.3 / 36.0 & 94.3\% / 93.0\% & 10.0\% / 10.2\% \\
\texttt{Compute}\,/\,Showdown sampling, $N{=}64$ & 4.9 / 132.0 & 98.4\% / 98.2\% & 4.8\% / 4.8\% \\
\texttt{Compute}\,/\,Showdown sampling, $N{=}256$ & 19.4 / 509.0 & 99.7\% / 99.4\% & 2.8\% / 2.6\% \\
\bottomrule
\end{tabular}
\caption{Recovering the outcome distribution of one joint action: \texttt{Search}-mode enumeration versus sampling, averaged over 500 random openings with random joint move actions.
Outcomes are distinct when they differ in any end-of-turn discrete features.
Coverage is the reference outcome mass the method observed, and misallocation is the total-variation distance from the reference weights.
A 32-branch cap is simultaneously faster and more accurate than sampling at any tested budget.}
\label{tab:outcome-coverage}
\end{table}

\subsection{Validation}
\label{sec:engine-results}
We test 500 random openings with random joint actions, resolved exhaustively in \texttt{Search} mode as a reference distribution, then replayed 256 times by Showdown and \texttt{Compute} mode (\cref{tab:outcome-coverage}).

\emph{Correctness}: the exhaustive enumeration should contain every outcome, so any sampled output landing outside it is a bug.
The first pass mismatched 6 of 500 turns and exposed four bugs, since fixed, and that first-pass ${\sim}99\%$ is the parity figure we quote (\cref{app:parity}).
Two regression suites guard it, more than 600 synthetic scenarios and over 500 recorded human ladder battles replayed turn by turn.

\emph{Performance}: as a plain simulator, \texttt{Compute} mode resolves a median turn in 0.08\,ms against Showdown's 2.0\,ms, a $26\times$ gap (single Node process; neither side's time includes transport).

\emph{Coverage}: enumeration yields a median of 9 and up to 18{,}176 weighted branches per joint action (\cref{tab:outcome-coverage}), and sampling's $1/\sqrt{N}$ error is widest where it matters. On the branchiest probe (960 distinct outcomes), 256 samples miss 17\% of the mass.

\section{Method}
\label{sec:method}

Our method began as an adaptation of ReBeL \citep{brown2020rebel}, which under the pressure of VGC grew into something close to Student of Games \citep{schmid2023student}: CFR over public belief states with a counterfactual value network at the leaves, the subgame grown incrementally under a compute budget, and the networks retrained from self-play.%
VGC adds two axes neither poker nor Go has, simultaneous joint actions and wide chance nodes, and is far more data-hungry per unit of search; our departures from SoG, summarized in \cref{tab:method-lineage}, are adaptations to those facts, not a new algorithm.

\subsection{Problem Formulation}
\label{sec:method-formulation}
An OTS VGC battle is a 2p0s game with simultaneous moves, imperfect information, and chance (\cref{sec:background-search}).
We write $1$ for the seat being solved for and $2$ for the opponent.

\paragraph{Worlds and beliefs.}
The \emph{public state} $s$ is everything derivable from the battle log (species and formes, HP in coarse buckets, status, stages, field).
A \emph{world} $w\in\mathcal W(s)$ is one hypothesis about which four of their six the opponent brought: $|\mathcal W(s_\text{preview})|=\binom{6}{4}=15$ at team preview, dropping to $\leq\binom{4}{2}=6$ once the lead pair is revealed.
A \emph{public belief state} $\beta=(s,b)$ pairs $s$ with the solver's posterior $b\in\Delta(\mathcal W(s))$, seeded by the team-preview solve (\cref{sec:method-solving}).
Each opponent \poke\ $j$ additionally hides a stat spread $\xi_j$, carried not as extra worlds but inside every world as a mixture over a candidate set $\mathcal S_j$ with weights $q_j\in\Delta(\mathcal S_j)$ (\cref{sec:method-spreads}).
Information is one-sided: seat 1's configuration is treated as known to both players, conceding the opponent our hidden spreads and reserves.
Conceding information makes the solved game harder for us, never easier, so $\sigma_1$ is conservative, though it never exploits what the opponent does not know.

\paragraph{Dynamics and strategies.}
At a decision node each seat picks $a_i\in\mathcal A_i$ simultaneously. The engine resolves the joint action $a=(a_1,a_2)$ in world $w$ to an enumerated weighted batch of successors, $\mathcal T(s,w,a)=\{(s'_k,w'_k,p_k)\}_k$ with $\textstyle\sum_k p_k=1$.
Seat 1 plays one strategy $\sigma_1\in\Delta(\mathcal A_1)$; seat 2, whose private information is $w$, plays one per world, $\sigma_2^w\in\Delta(\mathcal A_2(w))$.
The value network $v_\theta(\beta)\in\mathbb R^{|\mathcal W(s)|}$ returns one value per world, as in DeepStack's and SoG's counterfactual value networks \citep{moravcik2017deepstack,schmid2023student}; $z\in\{-1,0,+1\}$ is a realized game outcome.

\subsection{Subgame Solving}
\label{sec:method-solving}
We solve every decision the agent faces, and every decision node inside a subgame, as a Bayesian matrix game.
Chance first draws the world $w\sim b$, which seat 2 observes and seat 1 does not; seat 1 plays $\sigma_1$, seat 2 plays $\sigma_2^w$, and chance resolves the joint action through $\mathcal T$.
The per-world payoff is
\begin{equation}
U_w(a_1,a_2)=\sum_{(s',w',p)\in\mathcal T(s,w,a)} p\,V(\beta'),\qquad V(\beta_{\mathrm{leaf}})=\langle b,\,v_\theta(\beta_{\mathrm{leaf}})\rangle,
\label{eq:payoff}
\end{equation}
where $\beta'$ carries the belief the strategies above it induce, and the node's value is
\begin{equation}
V(\beta)=\max_{\sigma_1}\;\min_{\{\sigma_2^w\}_w}\;\sum_{w}b(w)\sum_{a_1,a_2}\sigma_1(a_1)\,\sigma_2^w(a_2)\,U_w(a_1,a_2),
\label{eq:bayes}
\end{equation}
solved by alternating-update linear CFR \citep{brown2019solving} with one regret table for seat 1 and one per world for seat 2; the average profile $\bar\sigma$ converges toward a Nash equilibrium of the subgame.
The four decision types of a battle differ only in menus and in who holds private information (catalogued in \cref{app:nodes}, sketched in \cref{fig:solve}); the same solver runs all of them.

\paragraph{Team preview.}
Both rosters are public and nothing is hidden yet, so \cref{eq:bayes} collapses to a plain matrix game over the $\binom64\binom42=90$ (bring, leads) options per side, each cell opening the battle at a known world and solved as a turn node under the same budgeted expansion (\cref{sec:method-expansion}).
The solve is used twice: $\bar\sigma_1$ is sampled for our bring, and $\bar\sigma_2$'s bring marginal, blended with a corpus bring prior (\cref{sec:method-spreads}) and conditioned on the revealed leads, becomes the initial belief $b_0$: this is where the hidden information of every later $\beta$ is created.

\paragraph{Turn.}
A turn node's menus are the joint action choices of the two active slots (\cref{sec:background-vgc}), on the order of $10^2$ per side, so both menus are shortlisted by policy weighting before the matrix is solved.
Our menu is cut to the top $k_1$ by the policy prior, with the root cap configurable up to the whole menu.
The opponent's joint options are enumerated per world from that world's bench, aligned across worlds by identity, scored by the belief-weighted prior $\sum_w b(w)\,\pi^w_2(a_2)$, and capped at $k_2$, with slices of the cap reserved for switch and gimmick joints so those lines survive a low prior (\cref{app:nodes}).

\paragraph{Switches.}
After end-of-turn faints both sides pick replacements blind over at most $2{\times}2$ menus per world.
A pivot move instead pauses the turn with one seat choosing a replacement, and the chooser optimizes against a posterior that hypothesizes the other side's unseen committed actions, weighted by the pre-turn average profile and resolved from the paused state (\cref{app:nodes}).

\subsection{Subgame Expansion}
\label{sec:method-expansion}
The subgame is grown incrementally, interleaving expansion with regret updates as in GT-CFR~\citep{schmid2023student}.
We choose the continuation to expand next by SoG's selection rule adapted to the matrix game, a walk that mixes a PUCT argmax with the current average policy.
From the root, sample $w\sim b$, then each seat's action from
\begin{equation}
\tfrac12\,\mathrm{onehot}\Big(\arg\max_{a}\;Q_i(a)+c\,\pi_i(a)\,\frac{\sqrt{N}}{1+N_a}\Big)+\tfrac12\,\bar\sigma_i,
\label{eq:puct}
\end{equation}
with $Q_i$ the seat's counterfactual action values, $\pi_i$ the policy prior, and $N,N_a$ visit counts, then sample the chance outcome from $\mathcal T$, descend, and admit the first unexpanded continuation.

Unlike GT-CFR, we budget compute rather than node count.
A turn continuation costs $1$ and a mid-turn or forced-switch decision $c_{\mathrm{sw}}\ll 1$, each scaled by belief width $|\mathcal W|^{\rho}$ since engine and network work grow with a node's worlds.
Expansion halts when admitted cost reaches the budget $B$ (or its wall-clock equivalent at deployment), and in practice it binds long before any depth would.

\subsection{Hidden Stat Spreads}
\label{sec:method-spreads}
We model each opponent \poke's hidden stat spread as a mixture over a small candidate menu.
The menu $\mathcal S_j$ and its prior $q_j^0$ come from a team corpus, conditioned on the nature that open sheets reveal, with a reserved slice of base-stat archetype spreads so an off-meta build stays representable.
We rely on the engine for inference: after each turn the completed joint action is re-branched per world and per candidate, and a candidate is reweighted by the mass the engine's own transition puts on the observed public outcome and move order, so damage rolls and speed order discriminate through the exact rules with no hand-derived formula to keep in sync.
Posteriors persist across turns and the games of a Bo3 series (\cref{app:spreads}).

\subsection{Encoding and Networks}
\label{sec:method-encoding}
We encode a position's publicly observable information as features, along with two belief summaries, the weighted quantiles of each opponent \poke's stat-spread mixture and the turn phase with ours and hypothesized opponent's queued commitments.
One transformer trunk feeds both heads.
The policy head reads out the prior $\pi$ over a seat's joint two-slot menu, scoring the same joint actions the solver's matrices are built from, and the value head the per-world vector $v_\theta(\beta)$.
\Cref{app:encoding} gives the exact token structure, and \cref{sec:exp-setup} the sizes.

\subsection{Training}
\label{sec:method-training}
We initialize our networks by Behavior Cloning on public Showdown replays, similar to VGC Bench~\citep{angliss2025vgcbench}.
Replays carry no world labels, so cloning trains the policy head and a \emph{scalar} value head.
The per-world head is minted from the clone by tiling the scalar head's output across bring slots, so every world opens at the clone's prediction and self-play differentiates them.

We then trained in rounds, where fleets of self-play games run the search at every decision, every solve is harvested into training rows, then the networks are retrained and the next round starts from them.
The value target for world slot $w$ of a row is: %
\begin{equation}
y_w=\begin{cases}(1-\lambda)\,V_w+\lambda z & w=w^\star \text{ and the row is grounded}\\ V_w & \text{otherwise,}\end{cases}
\label{eq:targets}
\end{equation}
where $V$ is the solved backup and $w^\star$ the realized world.
We have 3 kinds of rows we train with:

\paragraph{Grounding.}
Pure solver targets bootstraps through the leaf network, so a drifted network can confirm its own bias, while the realized outcome $z$ is unbiased but noisy.
\Cref{eq:targets} mixes them on the slot of the world played.
Counterfactual worlds and interior nodes keep pure solver targets.

\paragraph{Interior supervision.}
Each solve has already computed values for its interior nodes, so we also train on them: up to $K$ per decision, sampled $\propto$ reach under $\bar\sigma$, each row weighted by its normalized reach and discounted by how much of its backup priced through the leaf network rather than terminating in-tree (sampling and weighting details in \cref{app:introws}).

\paragraph{Hypothetical augmentation.}
Self-play under-visits lopsided and unusual positions, so we manufacture them.
A fixed number of times per game, the live battle is forked at a decision boundary, a small set of edits is applied (HP, status, field, stat stages, or a hidden stat spread redrawn from the full usage prior), and the edited game is played out by the same search, contributing rows grounded by its own outcome.
This is similar in spirit to SoG's re-solved leaf queries.

\section{Experiments}
\label{sec:experiments}

\subsection{Setup}
\label{sec:exp-setup}

\paragraph{Format and teams.}
All results are in \poke\ Champions, VGC 2026 Regulation M-B doubles under OTS.\footnote{The ability Illusion is too rare in VGC to support, so the agent forfeits the occasional game against it.}
Self-play uses single game rewards, while online Showdown ladder plays best-of-three (Bo3) series.\footnote{Showdown format \texttt{gen9championsvgc2026regmbbo3}.}
Self-play draws teams from 663 unique public team pastes scraped from the community-maintained VGCPastes repository \citep{vgcpastes}.
Evaluation uses a hand-picked pool of 16 teams spanning the game's primary archetypes (``eval16'').

\paragraph{Agent.}
The networks instantiate \cref{sec:method-encoding} with a 256-wide, 6-layer, 8-head trunk (5.4M parameters).
It is initialized by behavior cloning on 122{,}804 human ladder replays (2.56M decisions).
Search follows \cref{sec:method}: up to 15 worlds per solve, budgeted PUCT expansion, interior menus of $16{\times}16$ joint actions, the opponent's root menu capped at 16, and one hypothetical fork per game.
The expansion budget $B$ follows the ramp in \cref{sec:exp-selfplay}; \cref{app:config} lists the remaining constants.

\paragraph{Evaluation protocol.}
An \emph{eval16} cell plays two agents over ordered pairs of the 16 eval teams with a fixed seed schedule: 256 games (each matchup 1 game, 95\% Wilson interval $\pm6.1$pp), and for certain close calls pooled to 1024 games (each matchup 4 games, $\pm3.1$pp).
Checkpoints play argmax of $\bar\sigma$ so that they are compared at their means (eval opponents do not adapt, so a deterministic policy cannot be exploited there), while ladder play samples at $T{=}0.5$.
Three fixed opponents recur: \textsc{d1sw15}, our strongest earlier-generation agent (\cref{sec:app-lineage}); \textsc{greedy}, a material-heuristic depth-one belief search; and \textsc{$\pi$-only}, a network's policy head without search.

\paragraph{Compute.}
Self-play runs on ten cloud slices of 48 CPU game workers, each sharing one L40S inference server; training uses one A100-40G. The 32-round run below costs roughly 250 L40S-hours of self-play and ${\sim}15$ A100-hours of training; the ladder runs on one local RTX~4090.

\subsection{The Training Run}
\label{sec:exp-selfplay}

\begin{figure}[t]
\settoheight{\figstrengthht}{\includegraphics[width=0.54\linewidth]{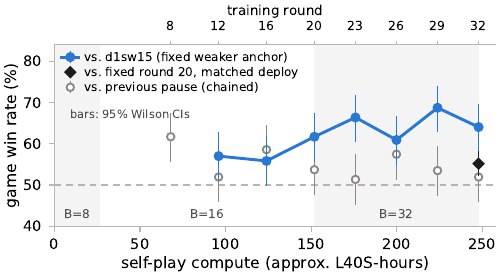}}%
\begin{minipage}[b]{0.54\linewidth}
\centering
\includegraphics[width=\linewidth]{figures/fig_training_strength}
\caption{Checkpoint strength by three references: the fixed older agent \textsc{d1sw15} (inflates as the lineage learns to exploit it), the previous pause checkpoint (leans high when chained), and fixed round 20 at matched deployment search (the read we use).}
\label{fig:training-strength}
\end{minipage}\hfill
\begin{minipage}[b]{0.42\linewidth}
\centering\scriptsize
\setlength{\tabcolsep}{2.5pt}
\renewcommand{\arraystretch}{1.05}
\begin{minipage}[c][\figstrengthht][c]{\linewidth}
\centering
\begin{tabular}{@{}lrlr@{}}
\toprule
Deployment variant & \multicolumn{2}{l}{Win\% [95\% CI]} & $p_{50}$ \\
\midrule
budget $B{=}128$ & 50.8 & [44.7, 56.8] & 2 \\
interior menus $24{\times}24$ & \textbf{42.6} & [36.7, 48.7] & 2 \\
\addlinespace
$B{=}64$, own 32 & 53.7 & [47.6, 59.8] & 2 \\
$B{=}128$, own 32 & 56.6 & [50.5, 62.6] & 3 \\
$B{=}128$, own 64 & 52.0 & [45.8, 58.0] & 3 \\
$B{=}256$, own 32 & 59.0 & [52.9, 64.8] & 3 \\
\addlinespace
$B{=}128$, own 32, iters 200 & \textbf{63.7} & [57.6, 69.3] & 3 \\
$B{=}256$, own 32, iters 200 & 59.6 & [53.5, 65.5] & 3 \\
\bottomrule
\end{tabular}
\end{minipage}
\captionof{table}{Round-32 network, variant vs.\ native deployment ($B{=}32$, 100 child CFR iterations), eval16 cells. Own $k$: child menus widened to $k$; iters: child CFR iterations; $p_{50}$: median solve depth in turns.}
\label{tab:deploy}
\end{minipage}
\end{figure}

The reported run is 32 rounds of the loop of \cref{sec:method}, ramping $B$ from 8 to 32 and games per round from 1k to 3k (schedule in \cref{app:config}). By the matched-lineage read of \cref{fig:training-strength}, the back half of the run bought $+5$pp over round 20.

\paragraph{Diminishing returns at $r{=}32$.}
Training progress has not stopped, but its price has collapsed.
Eight further rounds at $B{=}16$ end at parity with round 32 head-to-head (50.8\%, 1{,}024 games), and a $B{=}32$ continuation arm does no better (48.1\%). A $B{=}64$ extension with deepened child solves, at $1.6\times$ the cost per round, peaks at 54.5\% in its fourth round (1{,}024 games) but sits at 49--51\% in rounds three, six, and nine.
Round 32 remains the agent evaluated below.

The plateau leaves the network's early-game optimism in place. It shows on self-play roots, certified positions, and ladder solves, and neither further rounds nor deeper solves move it (\cref{sec:exp-value}).

\paragraph{Search carries the agent.}
\textsc{greedy} beats every raw policy head we trained (71\% pooled) and a PPO self-play policy built with the VGC-Bench recipe \citep{angliss2025vgcbench} at 60.5\% (\cref{sec:app-recipe}). Every search agent beats its own policy head, and our round-one network under search beats that PPO policy 73--27. Self-play widens the lead, as the behavior-cloned parent wins only 19.5\% against round 32 under the same search.
Almost every training-side lever lands within an eval16 cell's noise (\cref{sec:exp-ablations}). Compute and the encoder moved strength, not target construction, and the hidden-stat belief summary was worth $+8$--$12$pp (\cref{sec:app-lineage}).

\paragraph{Equilibrium and exploitability.}
Decoupled UCT \citep{shafiei2009uct} computes no equilibrium. Run in place of CFR at every solve, with the same network, tree, and menus and ten times the iterations, it plays the native agent at parity (51.7\%, 2{,}048 games).
An informed local best response \citep{lisy2017lbr} that sees the agent's strategy and the true world separates the two. It wins 83.6\% against the decoupled-UCT mix and 49.6\% against the CFR mix (\cref{sec:app-eq}).
Search carries the strength, and only an adapting opponent tests the equilibrium.

\subsection{Scaling Search at Deployment}
\label{sec:exp-deploy}

Each cell of \cref{tab:deploy} plays the round-32 checkpoint against itself, one side at the native training shape and one at a variant.
Three regularities organize the table. \emph{Budget alone saturates}: the PUCT walk descends through the product of two seat policies and a chance factor, so depth grows only logarithmically in budget, and the extra spend buys width the value network cannot use.
\emph{Width alone is harmful}: wider interior menus spread the fixed child iterations over a $2.25\times$ cell area and dilute per-cell CFR convergence.
\emph{What pays is deepening and sharpening child solves}: budget times a wider \emph{own} child menu converts spend into depth (solves reaching $\ge$3 turns: 18\% $\to$ 56\%), and doubling child CFR iterations on that shape is the largest single lever.
The two stack, though the shared anchor hides it. The $B{=}256$ combination reads worse than its $B{=}128$ parent yet beats it head-to-head (56.6\%), so we settle close calls by direct play.
The winning shapes cost 4--6$\times$ the native wall clock.

\subsection{Human Evaluation: Showdown Ladder}
\label{sec:exp-ladder}

The final agent played two 75-set runs on the live Showdown Bo3 ladder, each from a fresh account and one set at a time, against a field averaging ${\sim}1320$ Elo (max 1620).
One run used its native training shape (median 2.7\,s per decision), the other the head-to-head winner of \cref{tab:deploy} ($B{=}256$, own 32, iters 200) under a window-filling time budget (median 23\,s).
The native run won \textbf{44--31 sets (58.7\%)}, taking a fresh account from 1000 Elo to a peak of 1492, briefly inside the format's top 500.
The scaled run won \textbf{45--30 sets (60.0\%)}.%
\ifdefined\arxivbuild
\footnote{Both accounts' rating pages are public: \url{https://pokemonshowdown.com/users/tamagotchai07} (1376 Elo, GXE 73.7) and \url{https://pokemonshowdown.com/users/tamagotchai08} (1397 Elo, GXE 71.9), ratings as of the end of each run.}
\fi
The game records are 95--84 (53.1\%) and 98--77 (56.0\%), and the pooled sets are 89--61 (95\% Wilson interval 51--67\%), though a ladder's baseline is the field's rating, not 50\%. %
The eval16 builds are mid-tier (median Bradley--Terry rank 217 of 648), and a third run with corpus-top teams won 56.3\% of 142 sets (\cref{sec:exp-teams}), so stronger builds did not help.

The peak is not the level.
The native run fell back out of the top 500 within ten sets, and across both runs the accounts kept reverting to a band of \textbf{1350--1400 Elo}, with GXE (Showdown's estimated win rate against a random ladder opponent) of 72--74, which we report as the agent's strength.
The scaled shape, 59.6\% against the native shape internally (\cref{tab:deploy}), shows no noticeable ladder edge, $+2.9$pp of games over 354 and a lower GXE (71.9 vs.\ 73.7).
Win rate does not fall across a set, and in 83 opt-in private challenges (52 on challengers' own teams) the agent won 64\% of games (\cref{sec:exp-humans}).

The recorded solves show that the opponent model's shortlist contains the opponent's realized \emph{joint} action on 64\% of observable turns against 97\% per-slot marginal coverage, so what the 16-joint menu misses is correlations between the two slots, not action vocabulary. Widening the menu at matched cost buys no strength (\cref{sec:exp-ablations}).
And the solved root value is better calibrated than the network that seeds it (Brier 0.171 vs.\ 0.214), sharpening from 0.22 on turns 1--3 to 0.10 from turn 8, the defect of \cref{sec:exp-selfplay} observed live (\cref{sec:exp-value}).

\section{Related Works}
\label{sec:related}

The equilibrium-search lineage itself is covered in \cref{sec:background-search}. From the AlphaGo lineage \citep{silver2016alphago,silver2018alphazero,schrittwieser2020muzero} we borrow, as SoG did, the outer loop in which search produces training targets and retrained networks strengthen the next round of search.

\paragraph{\poke\ agents.}
Early singles agents were scripted or searched hand-built engines \citep{technicalmachine,lee2017showdownai}.
The closest system to ours is Foul Play \citep{foulplay}, a singles ladder bot whose Rust engine also branches a move pair into weighted outcomes.
It runs an independent MCTS per sampled opponent team with a handcrafted evaluation, whereas we solve an equilibrium over a belief state and train value networks by self-play.
Our engine is also much more sophisticated, carrying stat and damage uncertainty as in-state distributions forked only where a condition depends on them, whereas Foul Play fixes one spread per search and uses expected values.
Recent agents are LLM-driven or learned from data: PokeLLMon \citep{hu2024pokellmon} prompts an LLM with battle state, PokeChamp \citep{karten2025pokechamp} adds minimax lookahead over LLM-sampled actions, Metamon \citep{grigsby2025metamon} reaches human-level singles ladder play by offline RL on human replays and self-play data, and VGC-Bench \citep{angliss2025vgcbench} provides a doubles benchmark with behavior-cloning, RL, and LLM baselines.
None computes equilibria at decision time: PokeChamp takes $\arg\max_{a}\min_{b}$ over sampled shortlists, Foul Play's MCTS selects each side's action by independent UCB1, and the rest act directly from a policy.
All but VGC-Bench target singles, and none is rated on the doubles ladder.

\paragraph{Simultaneous-move games.}
DeepNash \citep{perolat2022stratego} reached expert-level Stratego, a turn-based imperfect-information game with a hidden initial deployment, model-free via regularized Nash dynamics with no search at decision time; whether search pays its way in games of this shape is the question we take up.
Simultaneous moves also break standard MCTS: decoupled UCT selection, Foul Play's rule, can converge to exploitable strategies \citep{shafiei2009uct}, which SM-MCTS variants with regret-matching selection repair \citep{lisy2013convergence,tak2014variants}.
We treat simultaneity, hidden information, and chance by one mechanism, the CFR solve over belief subgames, and the concern resurfaces only in the expansion order (\cref{sec:method-expansion}).

\ifdefined\arxivbuild
\section{Limitations and Conclusion}
\label{sec:limitations}

\paragraph{Closed team sheets.}
Under closed sheets the hidden space grows from stat spreads to entire builds (moves, items, abilities, and spreads) jointly, and inferring sets from revealed events at that scale is unsolved in our framework.
Naively, some amount of Bayesian inference should work.

\paragraph{Team building and generalization.}
Team construction is out of scope, and generalization across diverse opponent archetypes is only partially exercised by online play.
Self-play does already rate teams (the Bradley--Terry fit of \cref{sec:exp-teams}), a signal a future team builder could search over.

\paragraph{Compute regime.}
The reported run, a single seed, cost roughly 250 L40S-hours of self-play and ${\sim}15$ A100-hours of retraining (\cref{sec:exp-setup}), a fleet of ten L40S GPUs at any moment.
That is ordinary for \poke\ agents, where Metamon and VGC-Bench each train on one 8-GPU machine \citep{grigsby2025metamon,angliss2025vgcbench}, but two to three orders of magnitude below the milestones our method descends from: ReBeL generated its poker data on 720 V100s \citep{brown2020rebel}, DeepNash trained on 1{,}024 TPU nodes \citep{perolat2022stratego}, and Student of Games reports TPU use comparable to its AlphaZero baseline, which ran 3{,}500 concurrent TPU actors \citep{schmid2023student}.
The diminishing returns of \cref{sec:exp-selfplay} are therefore a statement about this scale, not the recipe's ceiling: the value error that persists is the kind more data and capacity plausibly move.
We also never ablated the behavior-cloning initialization against a from-scratch lineage.

\paragraph{Value calibration.}
The value network's early-game optimism is the one defect our training did not move.
We suspect the domain's stochasticity is why: a realized outcome is a noisy label drawn from hundreds of possible outcomes per turn, so value learning needs far more games per unit of signal than in deterministic games.
\Cref{sec:exp-value} catalogues the evidence.

\paragraph{Conclusion.}
VGC breaks the tractability assumptions of the equilibrium-search lineage while remaining exactly simulable, which makes it a natural frontier domain for decision-time search.
We find that equilibrium search does scale to a game with simultaneous moves, wide chance nodes, and hidden information at once.
With an enumerating engine and a budgeted solver, the recipe that mastered poker holds 1350--1400 Elo against human VGC players.

\else
\section{Limitations and Conclusion}
\label{sec:limitations}

\paragraph{Limitations.}
Under closed team sheets the hidden space grows from stat spreads to entire builds, and inferring those from revealed events is unsolved in our framework.
Team construction is out of scope.
The reported run, a single seed, cost about 250 L40S-hours (\cref{sec:exp-setup}), two to three orders of magnitude below the milestones our method descends from \citep{brown2020rebel,perolat2022stratego,schmid2023student}. The diminishing returns of \cref{sec:exp-selfplay} describe this scale, not the recipe's ceiling.
Early-game value optimism is the one defect training did not move. We suspect label noise, as a realized outcome is one draw from hundreds of branches per turn (\cref{sec:exp-value}).

\paragraph{Conclusion.}
Equilibrium search does scale to simultaneous moves, wide chance nodes, and hidden information at once.
With an enumerating engine and a budgeted solver, the recipe that mastered poker holds 1350--1400 Elo against human VGC players.

\fi

\ifdefined\arxivbuild
\section*{Acknowledgements}

We thank the \poke\ Showdown project; this work would not have been possible without it.
We thank Modal for \$500 in compute credits.
We also thank Claude (Anthropic), which assisted with both development and the drafting and editing of this report.

\fi

\section*{Ethics Statement}
\phantomsection\label{sec:ethics}
\paragraph{Play against humans.}
The ladder results of \cref{sec:exp-ladder} come from games on the public Showdown ladder against anonymous opponents, who were not told that they faced a program.
Between July and September 2026 the agent played under two thousand ladder games across development and evaluation, 681 of them in the three reported runs.
We did not seek advance authorization for this play and relied on three grounds.
First, Showdown's published rules place no restriction on automated play.
Second, in a December 2025 policy review of ladder bots \citep{smogon2025ladderbots}, Showdown's stated concern was that they skew the usage statistics behind Smogon's tiers (fan-community-regulated formats), which does not apply to VGC, whose rules are set by the game's publisher.
That review closed in April 2026 with technical limits on ladder bots rather than a rule change, and our runs came after it.
Third, prior agents were evaluated on the same Showdown ladder \citep{hu2024pokellmon,karten2025pokechamp,grigsby2025metamon,foulplay}.
Each reported run used a fresh account and played one set at a time.
The agent kept within the format's clock (it never lost a game on time in those runs) and forfeited rather than stalled the games it could not play (\cref{sec:exp-setup}).
Opponents' usernames appear in our logs but are not reported, and will be removed from any released logs.
Every participant in the private challenge matches of \cref{sec:exp-humans} chose to challenge the agent and received, after each game, a recap of the agent's solved policy and its estimate of theirs.
\paragraph{Contact with Showdown.}
In August 2026 we opened a support ticket with Showdown's staff, describing the agent and its ladder rating and asking to discuss releasing its code and models.
Staff replied that they did not oppose the release but would likely revise the site's policy on bot accounts in the near future.
The following week we emailed the staff address a draft of this paper, raising the concern that released code would make ladder bots easier to build and asking how they would prefer we release it, and we have not received a reply.
\paragraph{Release.}
Our main concern is proficient bots flooding the public ladder, so we intend to release PokaiEngine and PokaiTrainer with the live-ladder client removed and to withhold trained checkpoints while their format is in play.
Behavior cloning used publicly archived replays, and the usernames in them are not used in training.

\section*{Reproducibility Statement}
Every mechanic the engine implements is checked against \poke\ Showdown by the parity harness of \cref{sec:engine} and \cref{app:parity}, and every number in \cref{sec:experiments} is computed from logged games by the evaluation scripts. The hyperparameters, staging schedule, and run scale are listed in \cref{app:config}.
We intend to release PokaiEngine and PokaiTrainer once the live-ladder client has been removed from the shipped build.
Release is not conditioned on Showdown's approval, though staff have said they do not oppose it (\hyperref[sec:ethics]{Ethics Statement}).
Trained checkpoints will not be released while the format they were trained on is in play, and we will revisit this when the format rotates.
\ifdefined\arxivbuild
For a limited time after this preprint, the agent was also reachable for private friendly battles on Showdown: it challenged any user who messaged it, and never entered a ladder queue; see \url{https://x.com/waxhn/status/2093142369923412145}.
\fi

\section*{AI Use Statement}
In this work we used generative AI tools (Claude, Anthropic) for the following tasks with required disclosure: implementing methods, designing and providing feedback on experiments and evaluation methodology, proposing and refining hypotheses, cleaning and reformatting the replay dataset, and interpreting results.
In particular, AI served as a brainstorming partner in designing the equilibrium-search method, while the engine design is the authors' own.
We did not use them to generate synthetic data or to develop theoretical models, and the work involves no mathematical proofs, translation, or qualitative data analysis.
We also used them for tasks with recommended disclosure: writing code, suggesting experimental parameters, creating figures, drafting and editing the text, searching and summarizing related literature, formatting references, and brainstorming.

We have reviewed all AI-assisted work.
Code was reviewed by the authors and is exercised by the engine's parity harness against \poke\ Showdown (\cref{sec:engine}) and the project's test suites.
Every reported number was recomputed from logged game artifacts, and literature summaries and citations were checked against the original sources.
All text was reviewed and edited by the authors.
We take responsibility for the final content of this work, including text, claims, and artifacts produced with the aid of generative AI.

\bibliographystyle{iclr2027_conference}
\bibliography{references}

\clearpage
\appendix
\crefalias{section}{appendix}
\crefalias{subsection}{appendix}
\crefalias{subsubsection}{appendix}
\section{Engine Details}
\label{app:engine}

\subsection{System Architecture and Backends}
\label{app:architecture}

\begin{figure}[t]
\centering
\resizebox{\textwidth}{!}{\begin{tikzpicture}[
  font=\scriptsize, >={Stealth[length=1.8mm]},
  mod/.style={draw, rounded corners=2pt, fill=white, align=center, inner sep=3pt, minimum height=0.55cm},
  sub/.style={draw, rounded corners=1pt, fill=gray!6, align=center, inner sep=2pt, font=\tiny, minimum height=0.5cm, minimum width=3.2cm},
  crate/.style={draw, rounded corners=3pt, fill=white, inner sep=5pt},
  ctitle/.style={font=\scriptsize, inner sep=1pt},
  layer/.style={draw, rounded corners=4pt, fill=#1},
  ltitle/.style={font=\scriptsize\bfseries, inner sep=2pt},
  flow/.style={->, thick},
  data/.style={->, thick, blue!65!black},
  lbl/.style={font=\tiny, fill=white, inner sep=1pt, align=center},
]
\node[ctitle, anchor=west] (tsd) at (0.3,0) {\texttt{poke-showdown}\quad \tiny Showdown protocol adapter};
\node[sub, minimum width=6.7cm] (tracker) at (3.7,-0.75) {\texttt{BattleStateTracker} + \texttt{reconstruct}\\[-2pt]protocol lines $\to$ public snapshot $\to$ player-view \texttt{BattleState}};
\node[sub] (subp) at (2.0,-1.55) {\texttt{simulate-battle}\\[-2pt]Node subprocess};
\node[sub] (ws)   at (5.4,-1.55) {WebSocket client\\[-2pt]login, matchmaking, rooms};
\node[ctitle, anchor=west] (ten) at (7.85,0) {\texttt{poke-engine}\quad \tiny batched turn pipeline};
\node[sub, minimum width=2.5cm] (compute) at (8.95,-0.75)  {\texttt{Compute} mode\\[-2pt]one timeline};
\node[sub, minimum width=2.5cm] (search)  at (12.6,-0.75)  {\texttt{Search} mode\\[-2pt]fork, prune, merge};
\node[sub, minimum width=2.5cm] (bengine) at (8.95,-1.55)  {\texttt{BattleEngine}\\[-2pt]exact simulation};
\draw[flow] (compute) -- (bengine);
\node[ctitle, anchor=east] (tapi) at (10.25,-4.0) {\texttt{poke-api}: \texttt{BattleEnv}\quad \tiny one seat-view surface};
\node[sub, fill=white] (bsub)    at (2.0,-3.2)  {\texttt{ShowdownSubprocess}\\[-2pt]sim + tracker, two seats};
\node[sub, fill=white] (bonline) at (5.4,-3.2)  {\texttt{ShowdownOnline}\\[-2pt]socket + tracker, one seat};
\node[sub, fill=white, minimum width=2.6cm] (blocal) at (8.95,-3.2) {\texttt{Local}\\[-2pt]shared engine, two seats};
\node[mod, thick, fill=yellow!10, minimum width=4.2cm, minimum height=1.35cm] (branch) at (12.95,-3.35)
  {\texttt{branch()} API\\[1pt]\tiny hypothesized \texttt{BattleState} + joint action\\[-2pt]\tiny $\to$ weighted outcome set\\[-2pt]\tiny};
\coordinate (rtop) at (0,0.55);
\coordinate (rbot) at (0,-4.35);
\begin{scope}[on background layer]
  \node[layer=orange!6, fit=(rtop)(rbot)(tsd)(search)(bsub)(branch), inner xsep=8pt, inner ysep=4pt] (rust) {};
  \node[crate, fit=(tsd)(tracker)(subp)(ws)] (showdown) {};
  \node[crate, fit=(ten)(compute)(search)(bengine)] (engine) {};
  \node[crate, fill=gray!6, fit=(tapi)(bsub)(bonline)(blocal), inner ysep=6pt] (api) {};
\end{scope}
\node[ltitle, anchor=north west] at ([xshift=2pt]rust.north west) {PokaiEngine (Rust crates)};
\draw[data] (subp.south) -- (bsub.north);
\draw[data] (ws.south) -- (bonline.north);
\draw[data] (bengine.south) -- (blocal.north);
\draw[data] (search.south) -- (search.south |- branch.north);
\coordinate (pline) at (0,-5.0);
\draw[dashed, gray!70] (rust.west |- pline) -- (rust.east |- pline)
  node[pos=0.72, lbl, text=gray!60!black] {\tiny \texttt{pokedojo} (PyO3 bindings)};
\coordinate (ptop) at (0,-5.6);
\node[mod, anchor=west] (drivers) at (0.05,-6.55) {Drivers\\[-1pt]\tiny self-play fleets, parity tests, ladder};
\node[mod, minimum width=3.2cm] (agent) at (7.45,-6.55) {\texttt{Agent} interface\\[-1pt]\tiny team preview + turn actions};
\node[mod, anchor=east] (impls) at (15.05,-6.55) {Agents\\[-1pt]\tiny search player, BC policy, heuristics, LLM};
\draw[flow] (drivers) -- node[lbl, above=2.5pt] {\tiny state, legal} node[lbl, below=2.5pt] {\tiny actions} (agent);
\draw[flow, dashed] (impls) -- node[lbl, above=2.5pt] {\tiny implements} (agent);
\begin{scope}[on background layer]
  \node[layer=green!5, fit=(ptop)(drivers)(agent)(impls), inner xsep=8pt, inner ysep=6pt] (py) {};
\end{scope}
\node[ltitle, anchor=north west] at ([xshift=1.5cm]py.north west) {\texttt{pokebench} (Python)};
\draw[data, <->] ([xshift=-1.35cm]drivers.north) -- node[lbl, right, pos=0.55] {\tiny state, legal actions, submit} ([xshift=-1.35cm]drivers.north |- api.south);
\draw[data] (impls.north) -- node[lbl, left, pos=0.3] {\tiny \texttt{branch()} on hypothesized states} (impls.north |- branch.south);
\end{tikzpicture}}
\caption{System architecture.
The engine runs one rules code in \texttt{Compute} or \texttt{Search} mode, the latter exposed as the \texttt{branch()} API returning weighted outcome sets; the Showdown adapter reconstructs the same player-view \texttt{BattleState} the engine emits from protocol lines; and \texttt{BattleEnv} is one seat-view surface over all three backends.
Python agents implement one \texttt{Agent} interface and run unchanged on any backend; the search player additionally calls \texttt{branch()} on hypothesized states.}
\label{fig:architecture}
\end{figure}

\Cref{fig:architecture} details the stack behind the shared battle abstraction of \cref{sec:engine}: local engine play, an offline Showdown process, and the live Showdown ladder as interchangeable backends under one seat-view surface.

\subsection{Optimization Details}
\label{app:engine-opt}

\paragraph{Representing outcomes compactly.}
A \poke's HP is carried as a distribution over the damage rolls it has absorbed, and an opponent \poke's hidden stat spread can be carried as an in-state distribution under the simplifying assumption that spreads of different \poke\ are statistically independent.
In reality teammates are trained with one another in mind. We have not measured the strategic cost of ignoring that correlation, and accept it for the computation the compaction saves.
When a later event does depend on position within an in-state distribution, such as an HP threshold that triggers an item or ability, or a speed comparison against an uncertain stat, the engine splits the state there, partitioning the distribution into the branches that behave differently.

\paragraph{Sharing work across branches.}
Shared damage modifiers are computed once per batch, while per-state work (status-dependent pre-move checks, damage against each state's stages and spreads) iterates over members.
Batches are keyed on the attributes move resolution can share and that rarely diverge across a turn's forks (which \poke\ are alive, items, abilities, the resolved action sequence, the field state), while statuses, stat stages, and spread hypotheses stay per-member.

\subsection{Parity Measurement}
\label{app:parity}
Behavioral parity is measured three ways, and only the figure we quote is out of sample.
\emph{Random-opening probe.}
The 500 random openings of \cref{tab:outcome-coverage} were drawn once, and each was replayed 256 times by Showdown and checked against the exhaustive enumeration before any mismatch was examined.
This first pass mismatched 6 of 500 turns, traced to four engine bugs, and its rate is the ${\sim}99\%$ we report.
After the fixes the same probes pass 499 of 500, which we treat as a regression result rather than a parity estimate, since these probes found the bugs.
\emph{Scenario suite.}
More than 600 constructed positions, each isolating one mechanic, are played turn by turn on both simulators with their battle states asserted equal. They run in continuous integration, and all pass.
\emph{Human replays.}
527 recorded ladder battles from three regulations (199 each of Regulations M-A and M-B, 129 of Regulation I) are replayed turn by turn, with the engine driven by the recorded actions and its public state compared against the log after every turn.
The public state matches on all 3{,}452 turns. Of Showdown's per-turn events, 97\% are reproduced in order, and about 80\% of turns reproduce the full event sequence.
Both suites were used to find and fix bugs, so they bound regressions rather than estimate parity on unseen play.

\section{Method Details}
\label{app:method}

\subsection{Relation to ReBeL and Student of Games}
\label{app:method-lineage}

\Cref{tab:method-lineage} summarizes, component by component, where the method of \cref{sec:method} follows ReBeL and SoG and where the domain forced a departure.

\begin{table}[t]
\centering\small
\begin{tabular}{@{}p{0.27\linewidth}p{0.25\linewidth}p{0.42\linewidth}@{}}
\toprule
Component & Follows & Departure for VGC \\
\midrule
Belief states, CFR solve, CV net at leaves & ReBeL, SoG & simultaneous-move matrix nodes; exactly enumerated chance nodes (\cref{sec:method-solving}) \\
Incremental subgame growth & SoG (GT-CFR) & walk over a matrix game; admissions priced by compute cost, budget in cost units (\cref{sec:method-expansion}) \\
Value targets & SoG (solver values $+$ TD(1) rows) & per-row $\lambda$-mix, grounded on the realized world's slot only (\cref{sec:method-training}) \\
Extra supervision & SoG (re-solved leaf queries) & interior nodes of the same solve, reach-sampled and bootstrap-weighted \\
Off-line coverage & SoG (recursive queries) & hypothetical augmentation, grounded by play \\
\bottomrule
\end{tabular}
\caption{Where PokaiTrainer follows its lineage and where it departs.}
\label{tab:method-lineage}
\end{table}

\subsection{Decision Types of the Bayesian Matrix Game}
\label{app:nodes}
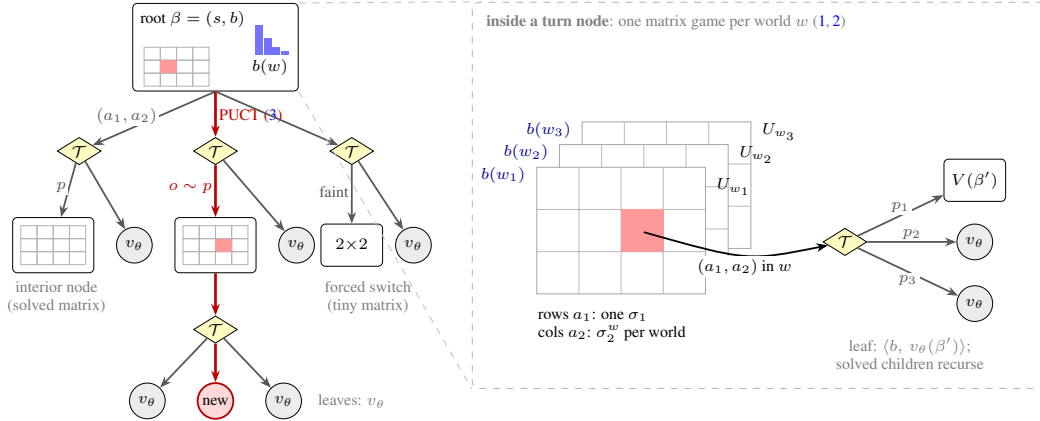
\begin{figure}[t]
\centering
\resizebox{\textwidth}{!}{\begin{tikzpicture}[
  font=\footnotesize, >={Stealth[length=1.8mm]},
  node/.style={draw, rounded corners=2pt, fill=white, inner sep=2pt, minimum width=1.25cm, minimum height=0.85cm},
  leaf/.style={draw, circle, fill=gray!15, inner sep=1pt, minimum size=0.55cm, font=\scriptsize},
  chance/.style={draw, fill=yellow!30, diamond, inner sep=1pt, aspect=1.6, font=\scriptsize},
  walk/.style={->, very thick, red!75!black},
  edge/.style={->, thick, gray!70!black},
  lbl/.style={font=\scriptsize, inner sep=1pt, fill=white},
]
\newcommand{\minimat}[3]{%
  \foreach \r in {0,1,2} \foreach \c in {0,1,2,3} {
    \draw[gray!60, line width=0.3pt] ($(#1,#2)+(-0.5+\c*0.25,0.3-\r*0.2)$) rectangle ++(0.25,-0.2);}
  \if\relax\detokenize{#3}\relax\else\fill[red!40] ($(#1,#2)+(-0.5+#3*0.25,0.3-0.2)$) rectangle ++(0.25,-0.2);\fi
}
\node[node, minimum width=2.55cm, minimum height=1.35cm] (root) at (3.0,0) {};
\node[font=\scriptsize, anchor=north west] at (root.north west) {root $\beta=(s,b)$};
\minimat{2.4}{-0.3}{1}
\foreach \i/\h in {0/0.45,1/0.25,2/0.12,3/0.06} {\fill[blue!55] ($(root.north east)+(-0.66+\i*0.13,-0.8)$) rectangle ++(0.11,\h);}
\node[font=\scriptsize] at ($(root.north east)+(-0.44,-0.98)$) {$b(w)$};
\node[chance] (ch1) at (0.9,-1.6) {$\mathcal T$};
\node[chance] (ch2) at (3.0,-1.6) {$\mathcal T$};
\node[chance] (ch3) at (5.1,-1.6) {$\mathcal T$};
\draw[edge] (root.south) -- node[lbl, pos=0.45, left] {$(a_1,a_2)$} (ch1);
\draw[walk] (root.south) -- node[lbl, pos=0.55, right] {PUCT (\ref{eq:puct})} (ch2);
\draw[edge] (root.south) -- (ch3);
\node[node] (n1) at (0.5,-3.05) {};
\minimat{0.5}{-3.05}{}
\node[leaf] (l1) at (1.75,-3.05) {$v_\theta$};
\draw[edge] (ch1) -- node[lbl, left, pos=0.45] {$p$} (n1);
\draw[edge] (ch1) -- (l1);
\node[node] (n2) at (3.0,-3.05) {};
\minimat{3.0}{-3.05}{2}
\node[leaf] (l2) at (4.25,-3.05) {$v_\theta$};
\draw[walk] (ch2) -- node[lbl, left, pos=0.45] {$o\sim p$} (n2);
\draw[edge] (ch2) -- (l2);
\node[chance] (ch4) at (3.0,-4.35) {$\mathcal T$};
\draw[walk] (n2) -- (ch4);
\node[leaf, fill=red!15, draw=red!70!black, thick] (new) at (3.0,-5.45) {new};
\node[leaf] (l3) at (1.95,-5.45) {$v_\theta$};
\node[leaf] (l4) at (4.05,-5.45) {$v_\theta$};
\draw[walk] (ch4) -- (new);
\draw[edge] (ch4) -- (l3);
\draw[edge] (ch4) -- (l4);
\node[node, minimum width=0.95cm, minimum height=0.65cm, font=\scriptsize] (fs) at (5.1,-3.05) {$2{\times}2$};
\node[leaf] (l5) at (6.05,-3.05) {$v_\theta$};
\draw[edge] (ch3) -- node[lbl, left, pos=0.45] {faint} (fs);
\draw[edge] (ch3) -- (l5);
\node[font=\scriptsize, gray, align=center] at (0.55,-3.85) {interior node\\(solved matrix)};
\node[font=\scriptsize, gray, align=center] at (5.35,-3.85) {forced switch\\(tiny matrix)};
\node[font=\scriptsize, gray, align=center] at (5.1,-5.45) {leaves: $v_\theta$};
\node[draw=gray!60, dashed, rounded corners=3pt, minimum width=8.9cm, minimum height=5.95cm, anchor=north west] (zoom) at (6.95,0.7) {};
\node[font=\scriptsize, anchor=north west, gray] at ($(zoom.north west)+(0.1,-0.08)$) {\textbf{inside a turn node}: one matrix game per world $w$ (\ref{eq:payoff},\,\ref{eq:bayes})};
\draw[gray!50, dashed] (root.north east) -- (zoom.north west);
\draw[gray!50, dashed] (root.south east) -- (zoom.south west);
\newcommand{\umat}[3]{%
  \draw[fill=#3, draw=gray!70] (#1,#2) rectangle ++(2.6,-1.95);
  \foreach \c in {1,2,3} \draw[gray!55, line width=0.35pt] (#1+\c*0.65,#2) -- ++(0,-1.95);
  \foreach \r in {1,2} \draw[gray!55, line width=0.35pt] (#1,#2-\r*0.65) -- ++(2.6,0);
}
\umat{8.65}{-1.15}{white}
\umat{8.3}{-1.5}{white}
\umat{7.95}{-1.85}{white}
\fill[red!40] (7.95+1.3,-1.85-0.65) rectangle ++(0.65,-0.65); %
\node[font=\scriptsize, anchor=west] at (11.3,-1.3) {$U_{w_3}$};
\node[font=\scriptsize, anchor=west] at (10.95,-1.65) {$U_{w_2}$};
\node[font=\scriptsize, anchor=west] at (10.6,-2.15) {$U_{w_1}$};
\node[font=\scriptsize, blue!60!black, anchor=east] at (8.6,-1.27) {$b(w_3)$};
\node[font=\scriptsize, blue!60!black, anchor=east] at (8.25,-1.62) {$b(w_2)$};
\node[font=\scriptsize, blue!60!black, anchor=east] at (7.9,-1.97) {$b(w_1)$};
\node[font=\scriptsize, anchor=north west, align=left] at (7.85,-3.95) {rows $a_1$: one $\sigma_1$\\cols $a_2$: $\sigma_2^{w}$ per world};
\node[chance] (zch) at (12.7,-3.0) {$\mathcal T$};
\draw[->, thick] (9.6,-2.85) .. controls (11.2,-3.3) .. node[lbl, below, pos=0.5] {$(a_1,a_2)$ in $w$} (zch);
\node[node, minimum width=0.95cm, minimum height=0.65cm, font=\scriptsize] (zs1) at (14.7,-2.05) {$V(\beta')$};
\node[leaf] (zs2) at (14.7,-3.0) {$v_\theta$};
\node[leaf] (zs3) at (14.7,-3.95) {$v_\theta$};
\draw[edge] (zch) -- node[lbl, above, pos=0.5] {$p_1$} (zs1);
\draw[edge] (zch) -- node[lbl, above, pos=0.5] {$p_2$} (zs2);
\draw[edge] (zch) -- node[lbl, below, pos=0.5] {$p_3$} (zs3);
\node[font=\scriptsize, gray, anchor=north, align=center] at (13.7,-4.35) {leaf: $\langle b,\,v_\theta(\beta')\rangle$;\\solved children recurse};
\end{tikzpicture}}
\caption{Anatomy of one budgeted solve.
Left: the subgame rooted at the public belief state $\beta=(s,b)$; every decision node holds a matrix game, chance nodes are the engine's exactly enumerated outcome batches $\mathcal T$, leaves are priced by $v_\theta$, and the red path is one PUCT expansion walk (\cref{eq:puct}) admitting a new continuation.
Right, zooming into a turn node: one payoff matrix $U_w$ per world, weighted by the belief $b$ and solved jointly by CFR (\cref{eq:payoff,eq:bayes}); each cell resolves through one \texttt{Search}-mode engine pass of $\mathcal T$, and its successors either recurse or read the leaf value $\langle b, v_\theta\rangle$.}
\label{fig:solve}
\end{figure}

\Cref{tab:nodes} catalogs the four decision types of \cref{sec:method-solving} as instances of the Bayesian matrix game, with the menus, the hidden information, and the resulting solve sizes of each; \cref{fig:solve} sketches the anatomy of one budgeted solve.
In the turn shortlist, a third of the opponent cap $k_2$ is reserved for switch-containing joints and a third of the remainder for joints that spend the gimmick.

\begin{table}[t]
\centering\small
\setlength{\tabcolsep}{4pt}
\begin{tabular}{@{}L{0.16\linewidth}L{0.08\linewidth}L{0.22\linewidth}L{0.22\linewidth}L{0.235\linewidth}@{}}
\toprule
Decision & Chooses & $\mathcal A_1$ / $\mathcal A_2$ & Hidden & Solve size \\
\midrule
Team preview & both & bring-4 $\times$ lead pair & none ($|\mathcal W|{=}1$) & $90\times90$ \\
Turn & both & move-or-switch per slot & $w\sim b$ & $|\mathcal A_1|\times k_2\times|\mathcal W|$ root; $k_1\times k_2\times|\mathcal W|$ interior \\
Forced switch (after faints) & both & replacement & $w\sim b$ & $\le 2\times2\times|\mathcal W|$ \\
Mid-turn switch (pivot) & one seat & replacement (one slot) & $w$, opponent's uncommitted actions $\mathcal H$ & $\le 2\times|\mathcal H|\times|\mathcal W|$ \\
\bottomrule
\end{tabular}
\caption{The decision types of a VGC battle as instances of the Bayesian matrix game (\cref{sec:method-solving}), with solve sizes as rows $\times$ columns $\times$ worlds.
$k_i$ is seat $i$'s menu after shortlisting by the policy prior (root solves keep our full menu; values in \cref{sec:exp-setup}); replacement menus are bounded by the two-\poke\ bench; $|\mathcal W|\le15$ before the leads are revealed and $\le6$ after.}
\label{tab:nodes}
\end{table}

\subsection{Observation Encoding}
\label{app:encoding}
\begin{itemize}[leftmargin=1.2em,itemsep=2pt,topsep=2pt]
\item \textbf{Tokens.}
A position is fourteen tokens: a summary token (a learned vector, which both heads read), our six \poke\ in roster order, the opponent's six in positional order (actives in field-slot order, then the revealed bench, then the sheet's unrevealed members), and one field token.
\item \textbf{\poke\ token.}
Embedded categorical ids (species, ability, item, the four moves, the two types, and the tera type) plus a 542-dimensional scalar block:
  \begin{itemize}[leftmargin=1.2em,itemsep=1pt,topsep=1pt]
  \item battle state: HP fraction, stat stages, PP, status, active flag and field slot, and 30 volatile conditions with their elapsed counters;
  \item stats: base stats, and calculated stats on our own side only (masked for the opponent, whose spreads are hidden);
  \item descriptors: one per move slot with the move's queued flag, and the ability and the item as currently held (a consumed or swapped item encodes as such);
  \item commitments: charging, recharging, semi-invulnerable, Protect streak, choice lock, and last move;
  \item revealed-ness: appeared, fraction of moves revealed, and brought / not brought / unknown;
  \item the two belief summaries of \cref{sec:method-encoding}: the phase-and-queue block (queue status, departing, switching-in slot, queued target and gimmick) and 27 stat-belief quantiles (mean, p10, p50, and p90 of each of the six stats under the spread mixture $q_j$, plus p10, p50, and p90 of the current HP fraction), which reduce to a point mass on our own side.
  \end{itemize}
\item \textbf{Field token.}
A 90-dimensional block: weather, terrain, rooms, gravity, both sides' side conditions with elapsed and remaining turns, the turn fraction, gimmick-used flags, the team-preview flag, and a one-hot of the decision phase (turn, forced switch, mid-turn switch).
\item \textbf{Belief as input.}
Each opponent token carries its marginal $P(\text{brought}\mid b)$, and the field token carries the full distribution over the $\binom64=15$ bring subsets of the open sheet in a canonical order, together with its effective sample size.
Nothing in the input asserts which world is true, since unrevealed opponent \poke\ are encoded as sheet members of unknown status, so one forward pass prices every world at once.
\item \textbf{Belief as output.}
The value head has fifteen units, one per bring subset in the same order: a world reads the unit of its bring, worlds that share a bring share a unit, and the leaf value $\langle b,v_\theta\rangle$ of \cref{eq:payoff} is the belief-weighted sum of the units.
\item \textbf{Trunk.}
A post-norm transformer with a learned positional embedding over the fourteen tokens; both heads read the summary token, the policy head scoring the joint two-slot menu.
\end{itemize}

\subsection{Interior-Row Sampling and Weighting}
\label{app:introws}

Up to $K$ interior rows per decision (\cref{sec:method-training}) are drawn $\propto$ reach under $\bar\sigma$ by Horvitz--Thompson systematic sampling: heavy nodes enter deterministically and the tail is sampled with inclusion weights, so $\mathbb E[\tilde r]$ reproduces the full capture's reach weighting.
Each sampled row is weighted by
\begin{equation}
\eta_{\mathrm{int}}=\tilde r\,\big(m+(1-m)\,\lambda_{\mathrm{int}}\big),
\label{eq:introw}
\end{equation}
where $\tilde r$ is the row's reach weight normalized within its solve and $m$ the fraction of its backup that terminated in-tree rather than at the leaf network, so the most network-priced backups are discounted.

\section{Hidden Stat Belief Machinery}
\label{app:spreads}
This appendix expands \cref{sec:method-spreads}.
\poke\ Champions allocates stat points on a $0$--$32$ per-stat scale with a $66$-point budget (in place of the $0$--$252$ effort values of the main series), so the investment grid per stat is small and the nature--spread pair is the whole hidden build under open sheets.
Spreads are assumed independent across \poke, so a world's spread belief is the product $\prod_j q_j$ of its per-\poke\ mixtures.

\paragraph{Prior fallback.}
A species' candidate menu comes first from the team-paste corpus (\cref{sec:method-spreads}): the top three spread rows observed with the species under its revealed nature (the unconditional marginal when the corpus never pairs the two).
Species the corpus lacks fall back to the per-species usage rows $\{(\xi_r,f_r)\}$ of \citet{championsbattledata}, scraped per season and format.
Usage reports nature and spread as independent marginals; since the nature is public under open sheets and the two are strongly dependent, each row is reweighted by nature compatibility $\kappa(\nu_j,\xi_r)\in\{1,\,0.35,\,0.01\}$: investment in the nature's lowered stat all but rules a candidate out, skipping its raised stat demotes it, and nothing is zeroed.
Raw rows are near-duplicates (the top rows of one species often differ by a point or two of filler), so they are greedily clustered: rows in order of $f_r\kappa$ join the first cluster whose representative is within $L_1$ distance $16$, pooling their mass, and the top three clusters become the candidates, spending slots on genuinely different builds (bulky versus fast) rather than variants of one.
In both the corpus and fallback paths, a fixed $\tau=0.15$ of the mass goes to base-stat archetype spreads, $q^0_j=(1-\tau)\,q^{\mathrm{head}}_j+\tau\,q^{\mathrm{tail}}_j$, deduplicated against the head candidates.
Under closed sheets the whole build is hidden and the menu comes from a Bayesian team-paste corpus model instead.

\paragraph{Outcome likelihood.}
After turn $t$ we have observed the joint action $a$ and the public outcome $o_t=\phi(s_t)$, where $\phi$ projects an engine state to its public part, refined by the observed move-order signature.
Writing $w[\xi_j]$ for world $w$ with \poke\ $j$'s mixture collapsed to the single candidate $\xi_j$, the likelihood of that candidate is the mass the engine's own transition puts on what was seen,
\begin{equation}
L_j(\xi_j)=\!\!\sum_{(s',w',p)\in\mathcal T(s_{t-1},\,w[\xi_j],\,a)}\!\! p\;\mathbf 1[\phi(s')=o_t],
\quad
q_j(\xi)\leftarrow \frac{q_j(\xi)\,\max\{L_j(\xi),\,\epsilon\max_{\xi'}L_j(\xi')\}}{\sum_{\xi''}(\cdot)},
\label{eq:spread-update}
\end{equation}
with $\epsilon=0.05$, and the world weight takes $P(w)\propto P(w)\sum_\xi q_j(\xi)L_j(\xi)$, the same numbers by linearity of $\mathcal T$ in the mixture.
The update discriminates only \emph{within} the menu, so its failure modes are menu failures: a candidate that cannot reproduce one outcome is demoted by the floor $\epsilon$ rather than eliminated, and a menu none of whose candidates reproduces the observed move order is handled by the contradiction ledger below.

\paragraph{Per-turn update.}
Three updates run each turn in order.
\emph{Reveal conditioning} zeroes worlds inconsistent with new hard evidence (a fifth revealed species not in a world's bring, a move or ability outside its config); item mismatches are priced softly because items change hands mid-battle.
\emph{Action conditioning} is the ReBeL update $P(w)\propto P(w)\,\bar\sigma^w_2(a_2)$ against our own last solve, floored so a surprising action dampens a world rather than killing it.
\emph{Outcome conditioning} is the engine step: the completed turn's joint action is re-branched per world, and the matched mass of the branch on the observed public outcome (HP quantized to 16 buckets, refined by the observed move-order signature when it is reproducible) is $P_w(o\mid a)$.
The same re-branch is run per spread candidate by materializing the world with one \poke's mixture restricted to that candidate; because the branch is linear in mixture weights, the per-candidate and per-world likelihoods are literally the same numbers, not two approximations.
Hidden opponent slots whose action was not observed are handled as a completion mixture weighted by the pre-turn $\bar\sigma$, capped in group count since each group is one branch.
Likelihoods are floored relative to the best candidate (a single observation moves a posterior at most $20\times$), worlds are evaluated heaviest-first up to a cap, and the spread posteriors live on the tracker across turns and across the games of a best-of-three set, where builds are locked.
Worlds are rebuilt from the current state when every world is contradicted or the effective sample size collapses while reveals are incomplete.

\paragraph{Contradiction ledger.}
Within-support evidence is absorbed by the posterior, but an observation that no candidate reproduces conditions nothing and was previously discarded: the ordering channel fell back to the public-only channel for everyone, because mixed-channel scoring would move belief toward refuted candidates.
The ledger catches exactly this signature, a public outcome that reproduces under some candidate while the move order reproduces under none, which distinguishes a speed-shaped contradiction from a degraded log.
It then probes rather than inverts: candidate spreads are synthesized along the species' speed-investment axis, substituted one at a time into a value-copy of the world, and scored through the same branch-and-match pipeline, in a coarse-to-fine sweep, so Trick Room, Tailwind, paralysis, stat stages and spread-dependent abilities are correct by construction.
The feasible hull is stored as a unary interval fact (opponent speed versus one of our own, whose speed is known) or a pair fact (opponent versus opponent).
A second producer reads the end-of-turn residual phase, where Showdown applies chip and healing per bracket in speed order, giving a uniform-context speed observation on most turns with two or more holders in a bracket.
Facts are kept as supersets of the truth (ties inclusive, rounding outward) so they compose by intersection and never need revisiting; dominated facts are dropped on insert and tightenings are recomputed by closure.
A binding fact marks the tracker dirty, and the next rebuild filters the candidate menu through the closed constraints and synthesizes a minimal-perturbation candidate per binding constraint, using the same recipe helpers the probes use so a probed candidate and a synthesized one can never disagree.
The ledger persists across the games of a set.
Known gaps: an order fact between two opponent \poke\ that are both off-menu is not resolved by marginal sweeps, and bulk and knock-out-timing axes are not built.
The ledger was off in every training run, where opponent spreads are drawn from the same corpus mixture the menu is built from, and on in all three ladder runs of \cref{sec:exp-ladder}, where hand-tuned human spreads outside that mixture are exactly the case it targets.

\section{Team Strength and Pool Effects}
\label{sec:exp-teams}

This appendix rates specific team builds within the 2026 Regulation M-B metagame.
It is written mostly for readers who play VGC; the findings will mean little without some familiarity with the format.

\Cref{tab:eval16} lists the eval16 pool of \cref{sec:exp-setup}: sixteen public team pastes from the VGCPastes repository \citep{vgcpastes}, hand-picked to span the meta's primary archetypes.

\begin{table}[t]
\centering\small
\begin{tabular}{@{}lrp{0.66\linewidth}@{}}
\toprule
Team & Elo & \poke \\
\midrule
\href{https://pokepast.es/dc43d16bba9a4dac}{\texttt{mb486}} & 1676 & \emph{Charizard}, Garchomp, Venusaur, Incineroar, Toxapex, Annihilape \\
\href{https://pokepast.es/cd3d3b91c38be1ca}{\texttt{mb355}} & 1579 & \emph{Charizard}, \emph{Floette-Eternal}, Kingambit, Whimsicott, Basculegion, Garchomp \\
\href{https://pokepast.es/460ce7877ff3669a}{\texttt{mb520}} & 1531 & \emph{Tyranitar}, \emph{Staraptor}, Excadrill, Gholdengo, Sinistcha, Milotic \\
\href{https://pokepast.es/7b256e7e2f3219ff}{\texttt{mb300}} & 1523 & \emph{Raichu}, Basculegion, Garchomp, Kingambit, Whimsicott, Ninetales-Alola \\
\href{https://pokepast.es/3fca28a8dcab073d}{\texttt{mb443}} & 1520 & \emph{Charizard}, \emph{Aerodactyl}, Garchomp, Sylveon, Farigiraf, Kingambit \\
\href{https://pokepast.es/9a08df9d69cc892f}{\texttt{mb210}} & 1516 & \emph{Froslass}, \emph{Scovillain}, Lycanroc-Dusk, Kingambit, Basculegion, Sneasler \\
\href{https://pokepast.es/fe091d570792318b}{\texttt{mb142}} & 1508 & \emph{Metagross}, \emph{Dragonite}, Ninetales-Alola, Garchomp, Sneasler, Kingambit \\
\href{https://pokepast.es/57bcaea77dce77ec}{\texttt{mb340}} & 1505 & \emph{Blastoise}, \emph{Delphox}, Sneasler, Kingambit, Incineroar, Sinistcha \\
\href{https://pokepast.es/caef1bb145b86b9b}{\texttt{mb366}} & 1484 & \emph{Charizard}, Garchomp, Gholdengo, Whimsicott, Incineroar, Basculegion \\
\href{https://pokepast.es/062ba26e30b09342}{\texttt{mb521}} & 1482 & \emph{Gengar}, \emph{Swampert}, Archaludon, Politoed, Vivillon, Incineroar \\
\href{https://pokepast.es/4f4d18cccd280c1a}{\texttt{mb524}} & 1473 & \emph{Charizard}, Grimmsnarl, Basculegion, Pelipper, Venusaur, Archaludon \\
\href{https://pokepast.es/79effda80d55d3f4}{\texttt{mb65}} & 1462 & \emph{Staraptor}, Grimmsnarl, Garchomp, Gholdengo, Sinistcha, Incineroar \\
\href{https://pokepast.es/d7a41e1b2dc484d5}{\texttt{mb166}} & 1452 & \emph{Swampert}, \emph{Floette-Eternal}, Pelipper, Grimmsnarl, Sinistcha, Archaludon \\
\href{https://pokepast.es/23538a4ad7d20e6b}{\texttt{mb244}} & 1447 & \emph{Froslass}, \emph{Blaziken}, Kingambit, Sinistcha, Basculegion, Sneasler \\
\href{https://pokepast.es/4488b90de17995c9}{\texttt{mb333}} & 1426 & \emph{Pyroar}, \emph{Blastoise}, Torkoal, Farigiraf, Sylveon, Venusaur \\
\href{https://pokepast.es/5d9021e5d4ca36cd}{\texttt{mb235}} & 1418 & \emph{Mawile}, Kangaskhan, Farigiraf, Torkoal, Staraptor, Vivillon \\
\bottomrule
\end{tabular}
\caption{The eval16 evaluation pool, sorted by strength.
Each team is a public paste from the VGCPastes repository \citep{vgcpastes} (linked; author credit on the paste page), identified by its corpus index.
Elo: Bradley--Terry fit over 2{,}940 internal head-to-head games among the 16 (mean 1500).
Italics mark \poke\ holding a Mega Stone.}
\label{tab:eval16}
\end{table}

Eval16 was chosen for archetype diversity, not strength, and it turns out to be a mid-tier pool.
Rating every build in the corpus with a Bradley--Terry fit over 18{,}000 self-play games (648 canonical builds from 797 pastes, median 48 games each), the 16 eval teams scatter from rank 3 to rank 459 of 648 (median 217), with four in the bottom half.
The orderings are also strongly pool-dependent: the corpus-wide ratings agree with the eval16 round-robin hierarchy only weakly (Spearman $+0.38$), and neither correlates with per-team ladder residuals at current sample sizes.
Being good against these 16 is substantially a matchup fact, not a strength fact.
Both consequences matter for the main text: internal win rates are compressed by mirror-adjacent matchups, and the ladder runs pilot mid-tier public builds against opponents free to bring the meta's best.
A ladder run on a corpus-top pool is reported at the end of this appendix.

To rate the strong end under conditions matching live play, the 32 highest-rated builds also played a full round-robin (496 unordered pairs $\times$ 3 games $=$ 1{,}488 games) at a deepened deployment shape of \cref{sec:exp-deploy} ($B{=}128$, widened 200-iteration child solves).
\Cref{tab:teamrr} shows the standings; three readings stand out.

\begin{table}[p]
\centering\small
\setlength{\tabcolsep}{4pt}
\renewcommand{\arraystretch}{0.92}%
\begin{tabular}{@{}rlrrp{0.56\linewidth}@{}}
\toprule
\# & Team & Win\% & cRk & \poke \\
\midrule
1 & \href{https://pokepast.es/597cbdbdcd0338da}{\texttt{mb1}} & 76.3 & 11 & \emph{Tyranitar}, \emph{Staraptor}, Excadrill, Basculegion, Sinistcha, Raichu \\
2 & \href{https://pokepast.es/89807d8bc9072b1d}{\texttt{mb194}} & 69.9 & 6 & \emph{Froslass}, Incineroar, Gholdengo, Basculegion, Kingambit, Arcanine-Hisui \\
3 & \href{https://pokepast.es/b3cf90d623c923c5}{\texttt{mb537}} & 68.8 & 19 & \emph{Charizard}, \emph{Mawile}, Whimsicott, Basculegion, Farigiraf, Garchomp \\
4 & \href{https://pokepast.es/03f74fec574e9b97}{\texttt{mb345}} & 65.6 & 13 & \emph{Charizard}, \emph{Garchomp}, Venusaur, Farigiraf, Incineroar, Sylveon \\
5 & \href{https://pokepast.es/126c2dd2034d9908}{\texttt{mb198}} & 64.5 & 20 & \emph{Swampert}, \emph{Floette-Eternal}, Whimsicott, Pelipper, Archaludon, Basculegion \\
6 & \href{https://pokepast.es/9e447be12c2fbc70}{\texttt{mb423}} & 63.4 & 7 & \emph{Raichu}, \emph{Tyranitar}, Meowscarada, Basculegion, Kleavor, Talonflame \\
7 & \href{https://pokepast.es/5b1fda1a561f194f}{\texttt{mb350}} & 62.4 & 15 & \emph{Raichu}, \emph{Floette-Eternal}, Whimsicott, Basculegion, Garchomp, Gholdengo \\
8 & \href{https://pokepast.es/48ca91ab5c767b46}{\texttt{mb200}} & 62.4 & 22 & \emph{Swampert}, \emph{Metagross}, Grimmsnarl, Pelipper, Archaludon, Sinistcha \\
9 & \href{https://pokepast.es/cf96601b598cf9dc}{\texttt{mb417}} & 62.4 & 5 & \emph{Charizard}, \emph{Aerodactyl}, Kingambit, Farigiraf, Garchomp, Sylveon \\
10 & \href{https://pokepast.es/dc43d16bba9a4dac}{\texttt{mb486}} & 61.8 & 3 & \emph{Charizard}, Garchomp, Venusaur, Incineroar, Toxapex, Annihilape \\
11 & \href{https://pokepast.es/1f79c6e87994defc}{\texttt{mb534}} & 61.3 & 17 & \emph{Charizard}, Whimsicott, Garchomp, Incineroar, Basculegion, Gholdengo \\
12 & \href{https://pokepast.es/f77ea22a4a5d1d15}{\texttt{mb695}} & 60.2 & 4 & \emph{Sceptile}, \emph{Tyranitar}, Rotom-Wash, Corviknight, Arcanine-Hisui, Houndstone \\
13 & \href{https://pokepast.es/f3fe5c118bfffece}{\texttt{mb675}} & 58.6 & 9 & \emph{Kangaskhan}, \emph{Charizard}, Incineroar, Hatterene, Farigiraf, Torkoal \\
14 & \href{https://pokepast.es/b94b2877b5dd4f7d}{\texttt{mb128}} & 58.1 & 14 & \emph{Kangaskhan}, Vileplume, Farigiraf, Torkoal, Basculegion, Kommo-o \\
15 & \href{https://pokepast.es/cdbd30e8545e0a99}{\texttt{mb326}} & 57.0 & 12 & \emph{Charizard}, Farigiraf, Venusaur, Garchomp, Incineroar, Sylveon \\
16 & \href{https://pokepast.es/18296ccd2e5e15dc}{\texttt{mb413}} & 57.0 & 10 & \emph{Delphox}, \emph{Blastoise}, Incineroar, Maushold-Four, Sinistcha, Sneasler \\
17 & \href{https://pokepast.es/e475bf46b1880293}{\texttt{mb383}} & 51.6 & 30 & \emph{Pyroar}, \emph{Floette-Eternal}, Vileplume, Torkoal, Incineroar, Farigiraf \\
18 & \href{https://pokepast.es/124511a5e19b5e92}{\texttt{mb354}} & 49.5 & 23 & \emph{Delphox}, \emph{Blastoise}, Incineroar, Sinistcha, Maushold, Sneasler \\
19 & \href{https://pokepast.es/6163ed4553860686}{\texttt{mb188}} & 45.2 & 1 & \emph{Metagross}, \emph{Swampert}, Sinistcha, Pelipper, Archaludon, Hydreigon \\
20 & \href{https://pokepast.es/af02713bf1e5db3f}{\texttt{mb348}} & 43.0 & 2 & \emph{Staraptor}, \emph{Delphox}, Gholdengo, Grimmsnarl, Basculegion, Sinistcha \\
21 & \href{https://pokepast.es/1390e1a2e1455810}{\texttt{mb727}} & 40.9 & 8 & \emph{Froslass}, \emph{Scovillain}, Kingambit, Basculegion, Sneasler, Lycanroc-Dusk \\
22 & \href{https://pokepast.es/3852c9127b5ed3b8}{\texttt{mb450}} & 39.2 & 16 & \emph{Charizard}, \emph{Floette-Eternal}, Whimsicott, Basculegion, Garchomp, Kingambit \\
23 & \href{https://pokepast.es/8eaa6c943c31955a}{\texttt{mb459}} & 39.2 & 21 & \emph{Gardevoir}, Maushold-Four, Kingambit, Garchomp, Talonflame, Basculegion \\
24 & \href{https://pokepast.es/73b10f321bb443d3}{\texttt{mb273}} & 36.6 & 25 & \emph{Charizard}, \emph{Floette-Eternal}, Sneasler, Sinistcha, Incineroar, Kingambit \\
25 & \href{https://pokepast.es/bd5a9e826aec9762}{\texttt{mb576}} & 36.6 & 29 & \emph{Dragonite}, \emph{Scovillain}, Gholdengo, Basculegion, Sylveon, Garchomp \\
26 & \href{https://pokepast.es/8631596549d4d8f3}{\texttt{mb580}} & 36.6 & 26 & \emph{Raichu}, \emph{Aerodactyl}, Sylveon, Kingambit, Basculegion, Sneasler \\
27 & \href{https://pokepast.es/8154b0b8ee732c62}{\texttt{mb341}} & 33.9 & 24 & \emph{Froslass}, Glaceon, Sneasler, Kingambit, Garchomp, Talonflame \\
28 & \href{https://pokepast.es/aef9b9811538653c}{\texttt{mb187}} & 31.2 & 28 & \emph{Floette-Eternal}, Whimsicott, Incineroar, Farigiraf, Garchomp, Basculegion \\
29 & \href{https://pokepast.es/4243181f96044698}{\texttt{mb603}} & 29.0 & 27 & \emph{Staraptor}, \emph{Delphox}, Garchomp, Whimsicott, Glimmora, Kingambit \\
30 & \href{https://pokepast.es/270173d4cbb96696}{\texttt{mb255}} & 26.9 & 31 & \emph{Charizard}, Bellibolt, Sableye, Garchomp, Basculegion, Sylveon \\
31 & \href{https://pokepast.es/407eef26d7044807}{\texttt{mb171}} & 26.3 & 18 & \emph{Eelektross}, Glimmora, Whimsicott, Basculegion, Garchomp, Farigiraf \\
32 & \href{https://pokepast.es/83dab8ecf3e23852}{\texttt{mb702}} & 24.7 & 32 & \emph{Gengar}, Swampert, Incineroar, Ninetales-Alola, Sneasler, Dragonite \\
\bottomrule
\end{tabular}
\caption{Full standings of the 32-team elite round-robin at the deepened deployment shape.
Win\% is over 93 games per team; cRk is the team's rank within the 32 by the corpus-wide fit.
Links and italics as in \cref{tab:eval16}.}
\label{tab:teamrr}
\end{table}

\paragraph{The strong end is the human meta's strong end.}
The top of \cref{tab:teamrr} is dominated by tournament-proven builds popularized by established competitors --- eval16's own \texttt{mb486}, tenth here, is the team of Shiliang Tang, a top VGC player and team builder --- and the winner is clear: the Mega Staraptor sand team \texttt{mb1} swept 16 of its 31 opponents.
The one entrant near the top a player would likely not have picked is \texttt{mb695}, a sand build around Mega Sceptile, rated fourth corpus-wide and holding 60.2\% against the elite pool.

\paragraph{Human-meta strength is a different axis.}
\texttt{mb702}, a Mega Gengar Yawn-trap team with an un-Mega'd Swampert, won a recent major tournament, so we forced it into the round-robin over its bottom-half corpus rating --- and it finished dead last.
Its gameplan monetizes surprise against human expectations, value the self-play-trained search neither finds nor falls for.
The corpus fit and the elite pool also disagree at the top in both directions (the cRk column): the corpus top two --- the \texttt{mb188} rain build and the dual-Mega Staraptor--Delphox ``StarFox'' team \texttt{mb348} --- both post losing records against the elite pool, while the winning sand team entered only 11th of the 32 by corpus rating.

\paragraph{Durable teams convert search.}
The corpus fit rates teams at self-play budgets $B\in\{32,64\}$; the round-robin plays a far deeper shape, and the rank moves between the two are systematic.
Durable builds rose --- \texttt{mb200}, Pelipper--Archaludon rain behind Grimmsnarl screens, from 22nd within the 32 to 8th, its screenless sibling \texttt{mb198} from 20th to 5th --- while the Mega Froslass plus Mega Scovillain hyper offense \texttt{mb727} fell from 8th to 21st. These moves confound search depth with the opponent pool, but a split of the corpus fit by budget, which holds the pool fixed, points the same way: the species gaining the most rating from $B{=}32$ to $B{=}64$ (Excadrill, Staraptor, Raichu) are three of the winning sand team's six.
Our reading is that bulkier structures give a deep solve more turns in which its edge compounds, while all-in offense has played out before the extra depth matters.

\paragraph{A corpus-top ladder run.}
We ran the ladder a third time at the scaled shape of \cref{sec:exp-ladder}, drawing from the top eight of \cref{tab:teamrr} (cut to the top six after 91 sets, dropping \texttt{mb194} and \texttt{mb423}), for 142 sets against a field averaging ${\sim}1310$ Elo.
The stronger pool bought nothing in aggregate: 80--62 sets (56.3\%) and 172--155 games (52.6\%), below both eval16 runs, with the round-robin winner \texttt{mb1} at 15--8 and the two teams cut at 5--8 and 5--7.
The run is also the paper's clearest picture of single-account noise: its first 81 sets went 52--29 (64.2\%) and reached 1415 Elo, and the remaining 61 went 28--33, including a 4--12 stretch that an audit of the build, the solve counters, and the opponents' ratings could attribute to nothing but variance.
Self-play rates a team within its pool; against the human field the rating is at best a weak prior.

\paragraph{Within-set adaptation.}
A best-of-three lets the same opponent face the agent up to three times, so exploitation within a set would show as a win rate that falls from game to game.
Over the three ladder runs the agent went 156--136 (53.4\%) in game one, 141--138 (50.5\%) in game two, and 68--42 (61.8\%) in game three.
Game one includes 13 sets in which the opponent walked out, and game three is played only after a split, but no decline appears.
Nor does game three rest on spread inferences carried over from earlier games. It went 39--30 when the opponent's roster had been fully seen in games one and two, and 29--12 when the opponent brought a \poke\ not yet seen, so the agent does at least as well when the opponent changes plan.

\paragraph{Private challenge matches.}
\label{sec:exp-humans}
After the ladder runs the agent was hosted as an opt-in practice partner. Any Showdown user who messaged the account could challenge it to a best-of-one or best-of-three, with the agent piloting either a house team or a team the user supplied as a paste.
In this mode the agent never entered a ladder queue.
After each game the agent posted a recap into the battle room, listing for every solved decision the root value of the solve, its own solved action mix, and the opponent mix it solved against, so a challenger could review why it played what it played.
Between 28 August and 20 September 2026, 13 users completed 83 matches (48 best-of-three and 35 best-of-one, at deployment budgets of 32, 64, and 256).
The agent won 57--26 by set and 98--56 by game (63.6\%).
In 52 matches it piloted one of 12 distinct user-supplied teams, seven of which match no self-play corpus team on all six species, and won 39--13 by set and 80--43 by game (65.0\%).
With house teams it went 18--13.
The sample is small and self-selected, and the users' ratings are unknown.
Two users account for 45 matches.
Against the heavier user (72 games, almost all with teams they supplied), the agent's win rate fell from 27 of 36 games in the first half to 21 of 36 in the second, while against the other (34 games) it rose from 10 of 17 to 12 of 17.
The data therefore neither shows nor rules out adaptation by a persistent opponent.

\section{Additional Experimental Results}
\label{app:experiments}

\subsection{Search and Training Configuration}
\label{app:config}

Defaults of the final run, beyond those stated in \cref{sec:exp-setup}: 300 CFR iterations at the root and 100 at interior nodes, 160 for the $90{\times}90$ preview; root own menu unrestricted; prune floor 2\% with at most 8 branches kept per joint action; belief-width exponent $\rho=1$ and switch-continuation cost $c_{\mathrm{sw}}=0.01$; interior rows capped at 96 per decision and weighted by walk visit share, with $\lambda_{\mathrm{int}}=0.15$ rising to 0.25 from round six; outcome-grounding mix $\lambda=0.5$; replay window shrinking from 5 to 2 rounds over the ramp.
The compute ramp: rounds 1--5 at $B{=}8$ and 1k games per round, 6--8 at $B{=}16$/2k, 9--20 at $B{=}16$/3k, and 21--32 at $B{=}32$/3k.

\subsection{The Reference Agent and Encoder Lineage}
\label{sec:app-lineage}

\paragraph{The \textsc{d1sw} reference agent.}
\textsc{d1sw} is a previous-tier agent trained for 15 rounds with a ``depth 1'' solve that exhaustively recurses forced-switch and mid-turn switch nodes but not actual turn nodes.
It is a hero run in the unflattering sense: it uses an older encoder with multiple bugs, an exhaustive search shape that costs $\sim$30\,s per decision, and a value head that passes only 26 of 170 direct-read tactical checks (\cref{sec:exp-value}).
It nevertheless beat every deeper, wider, and later agent we built for ten days of campaigns --- including its own search shape re-run on later encoder tiers --- so intermediate campaigns were reported against it round-matched, and its final checkpoint \textsc{d1sw15} remains the fixed foreign anchor of \cref{fig:training-strength}.
Round matching is not compute matching: \textsc{d1sw} rounds were 1--2k games at 30\,s per decision, where budgeted solves cost 10--23\,s.

\paragraph{Encoder tiers.}
The final run's encoder (\emph{v7}) is the earlier tier (\emph{v5}) plus the two belief summaries of \cref{sec:method-encoding}: v5 lacked the turn phase and queued commitments, so mid-turn leaves were priced as pre-turn states, and it left hidden stats to the belief alone, with no spread quantiles in the input.
Everything else in the agent (solver, targets, training) is indifferent to the tier.
The two summaries arrived in separate tiers, and only the second was measured in isolation: with every search and training knob held fixed, moving the recipe from the phase-only tier to v7 was worth $+8$--$12$pp against round-matched \textsc{d1sw} over its first five rounds (33--37\% $\to$ 42--46\%, 512-game cells), and the deeper 6-layer trunk added $\approx$6pp more by round five (49.9 vs.\ 43.8\%, one cell, $\approx2\sigma$).
The phase block's own contribution is confounded with search changes in the runs that introduced it, which still lost to round-matched \textsc{d1sw}.
Every cell here is a loss: at the round-matched budget of 8 the lineage trailed \textsc{d1sw} on every tier, and first reached parity only when the budget was raised to 32 (\cref{sec:exp-selfplay}).
The mid-game pair probes that motivated the spread quantiles are in \cref{sec:exp-value}.

\subsection{The Policy Recipe at Matched Training Compute}
\label{sec:app-recipe}
The \textsc{$\pi$-only} agents of \cref{sec:exp-selfplay} are the policy heads of our own networks.
To compare against the VGC-Bench recipe \citep{angliss2025vgcbench} on our own format, we also trained a policy by behavior cloning on the same public replays followed by PPO self-play, over the same Regulation M-B team corpus (361 teams at the time) and with a policy architecture carrying a 32-decision temporal context.
The PPO run made 2{,}000 iterations of 128 self-play games (256k games) on one RTX 4090 with 32 rollout workers, and its final policy reaches 39.5\% against \textsc{greedy} (101--155).
It beats the policy head of our round-one network 58--42, as a quarter-million PPO games should beat a head fit to one round of 1k searched games.
Under search the order reverses.
The same round-one network beats the PPO policy 73.0\% (187--69) at the native round-32 search shape and 74.2\% (190--66) at its own $B{=}8$ shape, and the behavior-cloned parent, which has seen no self-play at all, beats it 70.3\% (180--76).
The same parent under the same search loses 50--206 (19.5\%, Wilson [15.1, 24.8]) to the round-32 network. Search alone carries the recipe past PPO, and 32 rounds of self-play add thirty points on top.
Against \textsc{greedy} those three search cells are 54.3, 53.1, and 50.8\%.
Round one is the first and cheapest of the run's 32 rounds (\cref{app:config}), so the PPO side is not the compute-starved one, and training compute does not separate the recipes. Decision-time search does, before any self-play round has been paid for.
All cells are eval16, single seat, argmax, 256 games (${\pm}6$pp), and the comparison is to the recipe rather than to the released agents, whose team pool and regulation differ from ours.

\subsection{Training Ablations}
\label{sec:exp-ablations}

A caveat governs this subsection.
At our compute scale a campaign arm is two to seven rounds and a cell is 256--512 games, so the resolving power is $\pm4$--$6$pp, and almost every individual lever we tried lands inside that band.
Our working rule has therefore been to ablate in order to make sure a change is not worse, and then to keep the variant that makes more sense on its own terms (cheaper, simpler, better-covered gradients, better-calibrated), rather than to claim a win for it.
The arms below are reported in that spirit; \cref{tab:arms} collects the head-to-heads, cells against \textsc{d1sw} are round-matched (\cref{sec:app-lineage}), and win rates are over all games in a cell, so the rare game with no winner counts as a loss.

\paragraph{Expansion and budget.}
The budget binds on every walk: at $B\in\{8,16,32\}$ the spent cost reaches the budget on 100\% of walks and the early-stopping criterion never fires, while median walk depth grows 2$\to$3$\to$4.
Marginal units are cheap (per-walk wall 15.9$\to$23.6$\to$31.7\,s for $B=8\to16\to32$ on the local box), and budget spent at inference converts directly (\cref{sec:exp-deploy}; already visible mid-campaign, where a round-five network gained $+4.3$pp at $2\times$ its training budget).
Spent at data generation instead, doubling the budget did not move a plateaued lineage in two rounds (34.8/35.7\% vs.\ 35.5/33.6\%), foreshadowing the final run's budget-insensitive back half.
Replacing the best-first expansion heap by the PUCT walk held parity against the queue-mode sibling (48.8\% at round three, 52.0\% at round five) on roughly half the training rows per round and $-8$\% wall per decision, and was kept for that reason (the same arm cut hypothetical forks from three to one per game, so the two changes are confounded; the row saving is the walk's, which admits about a third as many forced-switch plans per decision, while the fork count was a wash on its own, \cref{tab:arms}); weighting interior rows by walk visit share instead of telescoped reach was a wash (50.0/52.3/52.3\% at rounds 1/3/5) and was kept because it un-kills the gradient on the 56\% of interior rows that had weight below $10^{-3}$.
Entry-point continuations (recursing the replacement and preview roots) won 52.0\% at round five and 57.0\% at round seven against their sibling; deploying the flags on the sibling's network alone gives 50.8\%, so the edge is in the training data.

\paragraph{Interior rows.}
On a round-one dataset (146k rows: 26k roots, 120k interior at a cap of six per decision) the interior share of the value loss is linear in $\lambda_{\mathrm{int}}$, from 10\% at $\lambda_{\mathrm{int}}{=}0$ to 43\% at 1; retraining across $\lambda_{\mathrm{int}}\in\{0,0.05,0.15,0.5,1\}$ is free up to 0.15 and costs $+0.003$--$0.005$ MSE and 0.07 of corr$(v,z)$ at 1, and by round six play is flat in $\lambda_{\mathrm{int}}$ (49.6/50.6/48.6\% against round-matched \textsc{d1sw} at 0.15/0.5/1.0, 512-game cells).
We use 0.15 rising to 0.25.
The one thing interior rows clearly did was qualitative: theirs was the first lineage whose self-play rounds did not erase the behavior-cloned parent's tactical discrimination (\cref{sec:exp-value}).
At the production cap of 96 rows the sampler captures 80--84\% of interior reach mass at $B\in\{8,16\}$; already at a cap of 24, interior rows are 94\% of a round's rows while carrying 23\% of its value weight.

\paragraph{Menus and solver iterations.}
Menu width never pays at matched cost in training-time cells either: on a $B{=}32$ round-five network, widening the opponent root cap from 16 to 32 is 47.3\% head-to-head against the native deploy, cutting our own root menu to 24 is 44.5\%, and uncapping both root menus is flat (53.1 vs.\ 51.2\% against \textsc{d1sw15}) at $1.84\times$ the cost; a shortlist probe on 59 self-play roots puts the knee at own 24 / opponent 16.
Root CFR iterations are a dead lever ($300\to2000$ moves the root value by 0.003, and $100\to400$ interior iterations flips no argmax --- though at the wider deployment shapes of \cref{tab:deploy}, where the same iterations spread over larger matrices, child iterations become the biggest lever), and CFR beats the fictitious-play solver of the ReBeL paper at equal wall clock (field win rate over eight rounds 49.5 vs.\ 46.0\%, 13.5k and 11.5k games against the same field).
Both solvers converge to an equilibrium, so this compares solvers. \Cref{sec:app-eq} tests a non-equilibrium baseline.
The one deploy-side menu lever that was unambiguous is shortlist \emph{ranking}: opponent menus ranked by the belief-weighted policy prior instead of by support count are worth $+16$pp on the same network (57.8 vs.\ 41.8\%, 512 games), and fixed the Protect-omission failure mode of the earliest ladder deployments, whose attack-biased shortlists had omitted Protect on 48\% of the turns it was legal.

\begin{table}[t]
\centering\small
\setlength{\tabcolsep}{4pt}
\begin{tabular}{@{}L{0.2\linewidth}L{0.27\linewidth}L{0.27\linewidth}L{0.2\linewidth}@{}}
\toprule
Arm & Change & Result & Verdict \\
\midrule
no grounding & $\lambda{=}0$, $\lambda_{\mathrm{int}}{=}1$ & 50.0\% (256/512) h2h, same deploy & wash at 2 rounds; calibration drifts \\
decoupled + zero-sum & $\theta'$ re-priced targets, moment loss, $\lambda{=}0$ & 42.6\% (109/256) vs sibling r3 & loses; dropped \\
hypothetical forks, 3/game & augmentation only & 48.8\% (125/256) compute-matched & wash per unit compute \\
hypothetical forks, 1/game + PUCT & & 52.0\% (133/256) on half the rows & kept \\
full-$z$ on hypo rows & $\lambda{=}1$ on augmented rows & 40.6\% (104/256) vs control & loses \\
entry continuations & replacement + preview roots & 52.0\% r5, 57.0\% r7 & kept \\
visit-weighted interior & $\tilde r$ from walk visits & 50.0/52.3/52.3\% & kept (gradient coverage) \\
\textsc{d1sw} shape on v7 & phased depth 1, switch-all & 43.0\% (110/256) r5 & loses \\
outcome-only head & re-init $v_\theta$, train on $z$ only & 49.2\% (126/256) vs r20 & parity; see \cref{sec:exp-value} \\
decoupled-UCT solver & solver swap at deploy, $10\times$ iterations, $+21$\% wall & 51.7\% (2{,}048 games) h2h; 85.5\% vs \textsc{greedy} & parity vs fixed field; exploitable (\cref{sec:app-eq}) \\
\bottomrule
\end{tabular}
\caption{Target and expansion ablations on the v7 deep line; each row is a head-to-head against its round-matched sibling unless noted.
Most levers are within a 256-game cell's $\pm6$pp of parity: the binding constraint is the value network, not the target construction.}
\label{tab:arms}
\end{table}

\paragraph{Targets.}
Removing outcome grounding costs nothing at two rounds but lets calibration drift ($+0.03$--$0.05$ on matched states, in-sample corr$(v,z)$ 0.64 vs.\ 0.78), so we kept it; re-pricing leaves with a frozen $\theta'$ plus a zero-sum moment loss at $\lambda{=}0$ loses outright; grounding augmented rows fully by their suffix outcome loses; hypothetical forks are a wash per unit compute and were kept at one per game for coverage.
\Cref{tab:arms} is the honest summary: at 2--5 rounds and 256 games, almost every lever lands within noise of parity.

\subsection{Equilibrium Quality and Exploitability}
\label{sec:app-eq}

\paragraph{Within the solved subgame.}
We instrumented every CFR call in 96 self-play games of the final agent on the eval16 pool (2{,}172 root solves, 138k child solves), recorded the per-world root payoff matrices after each solve, and computed the NashConv of the profile the agent plays from, in payoff units on $[-1,1]$.
An informed opponent's best response gains only $0.0004$ on average against the mix (median $0.0002$, p90 $0.0011$), with team preview at $0.0011$ and replacement games exact.
Cold Linear CFR on the same matrices reaches $0.041$ at 10 iterations, $0.0019$ at 100, and $0.0006$ at the deployed 300.
The solver is converged. This measurement does not bound what the subgame leaves out (menus, pruning, the leaf network).
The same matrices price the play temperature. The best response gains $0.117$ (p90 $0.31$) against the argmax the evaluation cells submit, $0.032$ against the ladder's $T{=}0.5$ draw, and $0.0004$ against the full mix.

\paragraph{The decoupled-UCT arm.}
The non-equilibrium baseline of \cref{sec:exp-selfplay} is decoupled UCT \citep{shafiei2009uct,tak2014variants}, the selection rule of Foul Play. Each seat runs UCB1 over its own actions and submits the most-visited one.
We swap it in for every matrix solve of the round-32 agent at its native shape, keeping the network, engine, menus, beliefs, and expansion walk, and give it ten times the iterations ($c{=}1$ on the $[-1,1]$ payoff scale). This costs 21\% more search wall time per game (148\,s against 122\,s).
It plays the CFR agent at parity, 51.7\% ([49.5, 53.8]) over 2{,}048 games, and beats \textsc{greedy} 85.5\% against CFR's 78.5\%.
On 400 captured root matrices its mix has NashConv $0.24$, against $0.0009$ for CFR at a tenth of the iterations, and the action it submits is exploitable by $0.23$, against $0.13$ for CFR's argmax.
Ten times the iterations halved its NashConv (from $0.43$ at 300), so more would not close the gap.
At these budgets its visit distribution is flat, with the most-visited row holding about a fifth of the root visits, so it is a weakly converged baseline rather than a greedy one.

\paragraph{An informed local best response.}
A local best response \citep{lisy2017lbr} bounds the exploitability of an agent whose best response is intractable.
At every decision the opponent is handed the agent's strategy and replies with the action that is best against it under a fixed local evaluation.
Ours is informed by the agent's own solved subgame and the true world.
It sits in the second seat and reads the distribution the agent actually draws from (a point mass under argmax, $\bar\sigma^{1/T}$ otherwise), never the realized draw.
It best-responds inside the agent's per-world payoff matrices, in the belief world that matches its own roster, and plays as the agent itself at team preview, at replacements, and on turns the agent decided without a solve.
Its win rate over 256 eval16 games lower-bounds the agent's exploitability (\cref{tab:lbr}), and 50\% means unexploited.
The CFR mix is unexploited, with an in-model gain of $0.0003$ per decision, and the per-decision gains under argmax and $T{=}0.5$ ($0.097$ and $0.027$) reproduce the within-subgame prices above in live play.
The decoupled-UCT agent plays fixed opponents at parity yet is exploited under every play mode, most of all under its own mix, because sampling from a flat non-equilibrium visit distribution plays many rows that a best response punishes.

\begin{table}[t]
\centering\small
\begin{tabular}{@{}llrlr@{}}
\toprule
Agent's solver & Agent's play & \multicolumn{2}{l}{Best-response win\% [95\% CI]} & Gain / decision \\
\midrule
CFR & argmax (eval cells) & 77.7 & [72.2, 82.4] & 0.097 \\
CFR & $T{=}0.5$ (ladder) & 52.3 & [46.2, 58.4] & 0.027 \\
CFR & $T{=}1$ (solved mix) & 49.6 & [43.5, 55.7] & 0.0003 \\
decoupled UCT & argmax & 74.6 & [68.9, 79.6] & 0.193 \\
decoupled UCT & $T{=}0.5$ & 72.7 & [66.9, 77.8] & 0.110 \\
decoupled UCT & $T{=}1$ & 83.6 & [78.6, 87.6] & 0.058 \\
\bottomrule
\end{tabular}
\caption{An informed local best response, seated second, against the round-32 agent under each solver and play mode (eval16, 256 games per cell). Gain is the best response's in-model advantage per decision over the agent's expectation, in payoff units on $[-1,1]$. The two solvers are at parity against fixed opponents (51.7\%, 2{,}048 games) and 34pp apart here.}
\label{tab:lbr}
\end{table}

\subsection{Value Network Quality}
\label{sec:exp-value}

\paragraph{Instrument.}
The tactics suite grades value networks against search-certified optima on hand-built Champions positions: 17 crafted positions (speed control, ability interactions, critical-hit lines, Perish Song, weather, gimmick timing, Protect reads) and 129 generated discrimination pairs, 248 positions in which one fact is flipped between a certified win and a certified loss (speed by raw stat, Trick Room, Tailwind, stages, Choice Scarf; bulk by matchup, screens, stages; burn, paralysis; weather damage and speed).
Values are certified by an exact solve whose leaves are re-filled with $0,\pm1$; fill-independence is the certificate.
Generated pairs are graded from the network's direct read at the root, since a depth-one solve rescues any network; we report \emph{root separation}, the mean read on certified wins minus certified losses, against an exact gap of about 2.0, and \emph{gap passes}, pairs the network separates by at least 0.7, over the 123 pairs whose certified gap has a direction.

\paragraph{The discrimination wall.}
Every checkpoint through mid-August separates wins from losses by only $+0.05$ to $+0.14$ (one $B{=}32$ arm reaches $+0.20$) and passes at most 3 of 123 gaps: the network gets the move right under search (crafted argmax 11--14/14, conversion walks 8--10/10) but does not see why.
The behavior-cloned parents are often the best direct readers, and earlier lineages' self-play rounds erased that knowledge (one parent passes 63/170 root checks, its self-play child 30/170); the interior-row lineage was the first whose rounds did not.
Only deep into the final run does the wall move: rounds 17--20 reach separations of $+0.16$ to $+0.23$ with 2--3 gap passes, and the $B{=}32$ back half pushes certified-loss hard passes to a campaign-high 25/120 by round 32.

\paragraph{Optimism.}
The final run prices certified losses positive in every round but the second (mean loss read up to $+0.75$): win-side root checks climb to 114/141 while loss-side checks stay at or below 13/120 through round 20, with the bias plateauing over rounds 11--15 ($+0.71$ to $+0.75$) and falling to $+0.10$ by round 32 before re-inflating in the saturated extension rounds.
The v5 family had the opposite sign.
The optimism is not a suite artifact: on ${\sim}36$k of its own self-play roots a round-two network reads $+0.12$ above the realized outcome on average, $+0.28$ on turns 1--2 falling to $+0.01$ past turn 7, the same phase profile the ladder solves show live (\cref{sec:exp-ladder}).
Re-initializing a late checkpoint's value head and training it on realized outcomes only (1.5M grounded rows, $\approx3$\% of the pool) recovers a head with the same play strength (49.2\% head-to-head, 61.3\% vs.\ \textsc{d1sw15}), better outcome correlation (0.72 vs.\ 0.69), and separation $+0.21$, yet still prices certified losses at $+0.22$: the blindness is a coverage problem of self-play, not a property of the solver targets or the BC initialization.

\paragraph{Where the error goes.}
On 33 late-endgame roots with certified value bands, budgeted solves ($B{=}16$) land inside the band for every network tested, including the worst-calibrated one (mean band distance 0.13, 79\% of direct reads outside): search absorbs leaf error where the horizon is short --- the mechanism behind the live calibration-by-phase profile of \cref{sec:exp-ladder}.
The depth-one shape leaks, and there the solve error tracks the network's oracle error (Pearson 0.40--0.86).
In mid-game pair probes the pre-v7 networks under-encode speed facts relative to \textsc{d1sw}'s (network-to-solve response ratio for Tailwind 0.15--0.27 vs.\ 0.53) and no network encodes screens, which a 16-unit solve proves worth $+0.11$ to $+0.19$; this probe is what motivated encoding raw spread quantiles in v7.

\end{document}